\pdfoutput=1
\documentclass[11pt]{article}

\usepackage[final]{acl}

\usepackage{times}
\usepackage{latexsym}

\usepackage[T1]{fontenc}
\usepackage[utf8]{inputenc}

\usepackage{microtype}

\usepackage{inconsolata}

\usepackage{graphicx}
\usepackage{multirow}
\usepackage{pifont}
\usepackage{booktabs}
\usepackage{hyperref}
\usepackage{makecell}
\usepackage{amsmath}
\usepackage{amsfonts}
\usepackage{enumitem}
\usepackage{amssymb}
\usepackage{multicol}
\usepackage[ruled,vlined]{algorithm2e}
\usepackage{algpseudocode}
\usepackage{soul}
\usepackage[most]{tcolorbox}
\tcbuselibrary{skins}
\tcbuselibrary{breakable}
\usepackage{bbding}
\usepackage{pifont}
\usepackage{scalerel}
\usepackage{colortbl}
\title{\scalerel*{\includegraphics{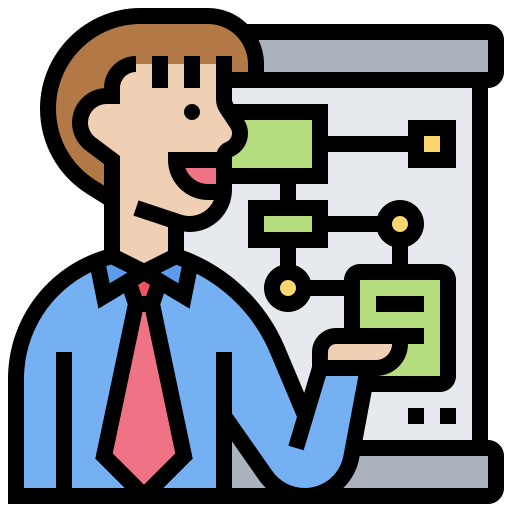}}{{\rule{3ex}{3ex}}}
MPO: Boosting LLM Agents with Meta Plan Optimization}

\author{Weimin Xiong$^{\heartsuit}$,
  Yifan Song$^{\heartsuit}$,
  Qingxiu Dong$^{\heartsuit}$,
  Bingchan Zhao$^{\clubsuit}$\\
  \textbf{Feifan Song}$^{\heartsuit}$,
  \textbf{Xun Wang}$^{\spadesuit}$,
  \textbf{Sujian Li}$^{\heartsuit}$\thanks{Corresponding Authors.}\\
  $^{\heartsuit}$Key Laboratory of Computational Linguistics, Ministry of Education, \\School of Computer Science, Peking University\quad\\
  $^{\clubsuit}$University of Washington\quad $^{\spadesuit}$Microsoft\quad\\
  \texttt{\{wmxiong, lisujian\}@pku.edu.cn} \\
  \vspace{-3mm}\\
 {\texttt{\url{https://github.com/WeiminXiong/MPO}}}
}

\definecolor{mintleaf}{RGB}{0, 184, 148}
\definecolor{dm-blue-500}{RGB}{0, 69, 177}
\definecolor{dm-purple-500}{RGB}{105,50,230}
\definecolor{mysilver}{RGB}{128,129,128}
\definecolor{my_green}{RGB}{0, 176, 80}
\definecolor{my_yellow}{RGB}{255,165,0}
\definecolor{my_red}{RGB}{255, 0, 0}
\definecolor{my_purple}{RGB}{126, 100, 158}
\definecolor{my_blue}{RGB}{49, 133, 155}
\definecolor{case_purple}{RGB}{160, 43, 147}
\definecolor{case_blue}{RGB}{15, 158, 213}

\newcommand{\planner}{meta planner}
\newcommand{\method}{MPO}

\begin{document}
\maketitle
\begin{abstract}
% Recent advancements in large language models (LLMs) have enabled LLM-based agents to successfully tackle complex interactive tasks. However, current planning methods in these agents often suffer from inefficiencies and planning hallucinations. 
Recent advancements in large language models (LLMs) have enabled LLM-based agents to successfully tackle interactive planning tasks. 
% However, despite their achievements, existing approaches often suffer from planning hallucinations and substantial training overhead.
However, despite their successes, existing approaches often suffer from planning hallucinations and require retraining for each new agent.
% In this paper, we introduce the \textbf{M}eta \textbf{P}lan \textbf{O}ptimization (\textbf{\method{}}) framework that enhances agent planning capabilities by incorporating explicit guidance. 
To address these challenges, we propose the \textbf{M}eta \textbf{P}lan \textbf{O}ptimization (\textbf{\method{}}) framework, which enhances agent planning capabilities by directly incorporating explicit guidance. 
Unlike previous methods that rely on complex knowledge, which either require significant human effort or lack quality assurance, \method{} leverages high-level general guidance through meta plans to assist agent planning and enables continuous optimization 
of the meta plans based on feedback from the agent's task execution. 
% We demonstrate the effectiveness of MPO through experiments on two representative benchmarks, ALFWorld and ScienceWorld, achieving performance improvements of up to 100\%. 
Our experiments conducted on three representative tasks demonstrate that \method{} significantly outperforms existing baselines.
% Further analysis shows that \method{} provides a plug-and-play solution that integrates seamlessly with existing agent frameworks, offering substantial gains in agent task completion efficiency and generalization to unseen scenarios.
% Moreover, our analysis indicates that \method{} provides a plug-and-play solution that enhances both task completion efficiency and generalization capabilities in previous unseen scenarios.
% Moreover, our analysis shows that \method{} provides a transferable solution readily applicable to new agent architectures, improving both task efficiency and generalization in previous unseen scenarios.
Moreover, our analysis shows that \method{} provides a portable solution that enhances both task completion efficiency and generalization capabilities across new agents and unseen scenarios.
\end{abstract}

\section{Introduction}
\begin{figure}
    \centering
    \includegraphics[width=\linewidth]{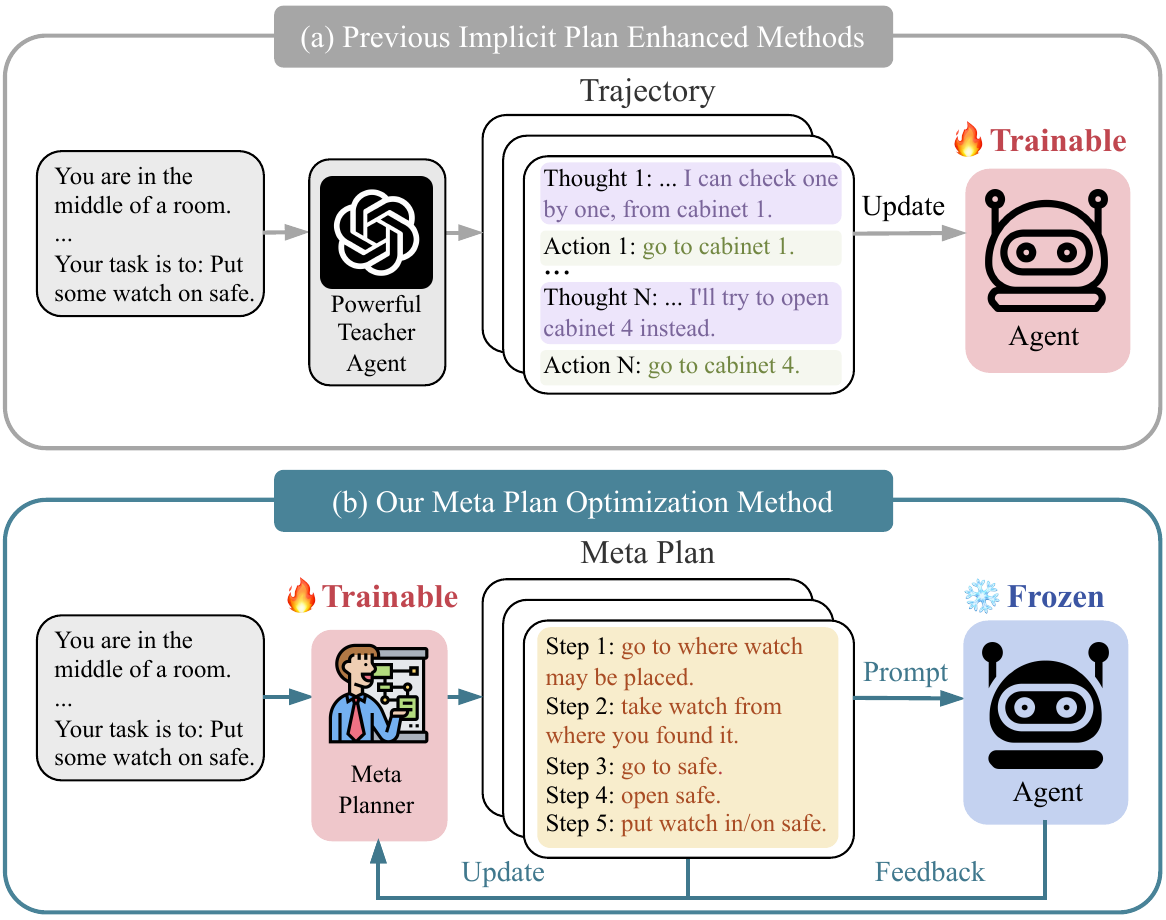}
    \caption{Unlike previous implicit plan enhancing methods that require agent parameter updates, our method incorporates meta plans into prompts for direct planning guidance and can improve them based on feedback.}
    \label{fig: teaser}
\end{figure}

Recent advancements in large language models~(LLMs)~\citep{achiam2023gpt, liu2024deepseek, yang2024qwen2} have enabled LLM-based agents to tackle complex multi-step tasks, including embodied housework~\citep{shridhar2020alfworld} and science experiments~\citep{wang2022scienceworld}. These tasks require sophisticated \textit{planning} abilities, as agents need to understand long-term dependencies~\citep{zhang2024survey}, reason about sequential actions, and adapt to dynamic environments~\citep{yao2022react}. 
The planning quality of these agents plays a crucial role in determining their overall performance.
Current mainstream LLM-based agents primarily develop their planning capabilities through implicit methods,
either directly leveraging the model's inner ability or fine-tuning from expert trajectories.
For example, ReAct~\citep{yao2022react} and Reflexion~\citep{shinn2024reflexion} perform planning \textit{on-the-fly} during task execution and are prone to getting lost due to planning hallucination~\citep{zhu2024knowagent}. 
% The works including AgentTuning~\citep{zeng2023agenttuning}, ETO~\citep{song2024trial}, and IPR~\citep{xiong2024watch}  employ trajectory tuning to enhance implicit planning capabilities and require retraining for each new agent, resulting in significant computational overhead (Figure \ref{fig: teaser}(a)).
The works including AgentTuning~\citep{zeng2023agenttuning}, Lumos~\citep{yin2023lumos}, and ETO~\citep{song2024trial}  employ trajectory tuning to enhance implicit planning capabilities and require retraining for each new agent, resulting in huge computational cost (Figure \ref{fig: teaser}(a)).

Beyond implicit planning, recent studies have explored the use of human knowledge to guide agents in task execution, capitalizing on the benefits of explicit guidance and low integration costs of such knowledge~\citep{zhu2024knowagent, DBLP:journals/corr/abs-2405-14205}. However, these approaches face significant challenges: they either require extensive manual efforts or struggle to ensure quality in the process of acquiring complex knowledge, potentially leading to inconsistent improvements in agent performance~\citep{wang2024astuteragovercomingimperfect}. 
To overcome these limitations, we propose to automatically generate a high-level abstract guidance, termed \textbf{Meta Plan}, which emulates human prior knowledge. Unlike previous implicit plans derived during execution, meta plans are decoupled from specific environmental details and complex agent trajectories, reducing the difficulty of knowledge acquisition.  Figure \ref{fig: teaser}(b) illustrates an abstract meta plan for the task "put some watch on safe". In contrast to the concrete plan in Figure \ref{fig: teaser}(a), the meta plan omits fine-grained details (e.g., cabinet 4).
%To further enhance meta plan quality, we design an iterative refinement mechanism that dynamically improves plans based on environmental feedback. This process mirrors how humans refine their strategies through experience, ensuring that meta plans evolve and adapt over time for optimal task execution.
To further enhance meta plan quality, we design a \textbf{M}eta \textbf{P}lan \textbf{O}ptimization (\textbf{MPO}) framework that iteratively improves plans based on environmental feedback. This process mirrors how humans refine their strategies through experience, ensuring that meta plans evolve over time for optimal task execution.

% In this paper, we propose the \textbf{M}eta \textbf{P}lan \textbf{O}ptimization~(\textbf{\method{}}), a framework consisting of a \planner{} and an act agent. 
%Furthermore, to control the quality of the meta plan, we propose  an optimization framework, \textbf{M}eta \textbf{P}lan \textbf{O}ptimization~(\textbf{\method{}}), that incorporates agent feedback to optimize the meta plan generator (\planner{}).
% Furthermore, to automatically improve the quality of the meta plan,
%Specifically, we propose  an optimization framework, \textbf{M}eta \textbf{P}lan \textbf{O}ptimization~(\textbf{\method{}}), which consists of a meta planner and an agent.
%Specifically, \textbf{\method{}} consists of a meta planner and a frozen  agent.
% Moreover, our proposed meta plan can be improved based on environmental feedback automatically. 
% Specifically, we propose an optimization framework, \textbf{M}eta \textbf{P}lan \textbf{O}ptimization~(\textbf{\method{}}), which consists of a meta planner and an agent.
%The meta planner is responsible for generating high-level meta plans, while the agent provides feedback on the execution to assess the quality of the inputted meta plans and help refine the meta planner.
The MPO framework comprises two key components: a meta planner and an agent.
The meta planner generates high-level meta plans, while the agent provides execution feedback to evaluate the quality of the input meta plans and guide meta planner refinement.
Initially, we collect meta plans from expert trajectories and cold-start the \planner{} through supervised fine-tuning.
To further optimize the \planner{}, we use Monte Carlo~(MC) sampling to estimate the task completion rate of the agent as feedback.
Specifically, given a task, the planner generates multiple meta plans through sampling. 
Then for each meta plan, the agent is also sampled to produce multiple execution trajectories, and the task completion rate is estimated accordingly. 
% After identifying contrastive meta plan pairs---those leading to the highest and lowest completion rates---we apply DPO~\citep{rafailov2024direct} to refine the \planner{} on these plan pairs.  
After identifying contrastive meta plan pairs---those yielding the highest and lowest task completion rates---we apply DPO~\citep{rafailov2024direct} to refine the \planner{} on these plan pairs.  
% Finally, the trained \planner{} can be detached from the MPO framework and function as a plug-and-play component, capable of generating high-quality meta plans for tasks in the target environment. This facilitates task completion for any new agent without incurring additional training costs.
% Finally, the trained \planner{} can be detached from the MPO framework and function as a plug-and-play component, capable of generating high-quality meta plans for tasks in the target environment. This facilitates task completion for any new agent without incurring additional training costs.
Finally, the trained \planner{} can be detached from the MPO framework and function as a portable component, capable of generating high-quality meta plans for tasks in the target environment. This facilitates task completion for any new agent without incurring additional training costs.

%The trained \planner{} serves as a plug-and-play component that can readily generate high-quality meta plans for tasks in the target environment, facilitating task completions for any new agent without additional training cost.

% We evaluate our approach on two representative benchmarks: ALFWorld~\citep{shridhar2020alfworld} for embodied household task and ScienceWorld~\citep{wang2022scienceworld} for textual science experiment task.
We evaluate our approach on three representative benchmarks: ALFWorld~\citep{shridhar2020alfworld} for embodied household tasks, ScienceWorld~\citep{wang2022scienceworld} for textual science experiment tasks and WebShop~\citep{yao2022webshop} for online web navigation tasks. 
% Across all test tasks, agents equipped with our \planner{} significantly outperform those without it, achieving performance improvements of up to 100\%. 
Across all test tasks, agents equipped with our \planner{} consistently outperform those without it, achieving at least a 5.6\% average improvement in performance.
% Additionally, the \planner{} is compatible with various existing agent frameworks.
% , including AgentTuning~\citep{zeng2023agenttuning} and ETO~\citep{song2024trial}. 
% When combined with these methods, out approach delivers even greater performance gains, which demonstrates its effectiveness in a larger application scope.
Additionally, the \planner{} is compatible with various agent training frameworks, and its meta plans can be directly inserted into task instructions. Combined with these methods, our approach yields even greater performance gains, demonstrating effectiveness in a larger application scope.
% Further analysis reveals that our generated meta plans substantially reduce the number of actions required for agents to complete tasks, thereby improving task completion efficiency. 
Further analysis reveals that our generated meta plans significantly increase the agent's average reward per action, thereby improving task completion efficiency.
% Moreover, we analyze why optimized meta plans enhance agent performance better than their unoptimized counterparts, as well as the impact of meta plan insertion positions on agent performance.

In summary, our contributions are as follows:

\begin{itemize}[leftmargin=*, nolistsep]
\setlength{\itemsep}{1mm}
% \item We introduce \textbf{\method{}} that explicitly incorporates meta plans and optimizes their generation. This advancement offers an innovative approach to enhancing agents' planning capabilities in a plug-and-play manner while remaining compatible with existing agent frameworks.
\item We introduce the \method{}, which leverages meta plan optimization to improve the performance of LLM agents.
% This progress offers an innovative approach to enhance agents' planning capabilities in a plug-and-play manner, while remaining its compatibility with existing frameworks.
% This progress offers an innovative approach to enhance agents' planning capabilities, while remaining its compatibility with existing frameworks.
This progress provides an innovative approach to explicitly enhance agents' planning capabilities while maintaining compatibility with previous agent training frameworks.
% \item Extensive experiments on two complex interactive tasks demonstrate that our method is applicable to a diverse set of agents and outperforms the baselines by a large margin.
\item Extensive experiments conducted on three representative benchmarks demonstrate that our method has significantly improved the performance of existing LLM agents.
% \item Further analysis demonstrates that incorporating the meta-plan generator reduces the number of actions required to complete a task, significantly improving the agent's task efficiency. Additionally, we qualitatively analyze the key factors that contribute to the performance improvements of the agent with the optimized meta plan.
\item Further analysis indicates that: (1) Our proposed method substantially boosts the agent's task completion efficiency; 
(2) A lightweight meta planner can guide more powerful agents in their planning.
and (3) \method{} increases the correctness, followability, and standardization of the meta plan.
\end{itemize}

\begin{figure*}
    \centering
    \includegraphics[width=\linewidth]{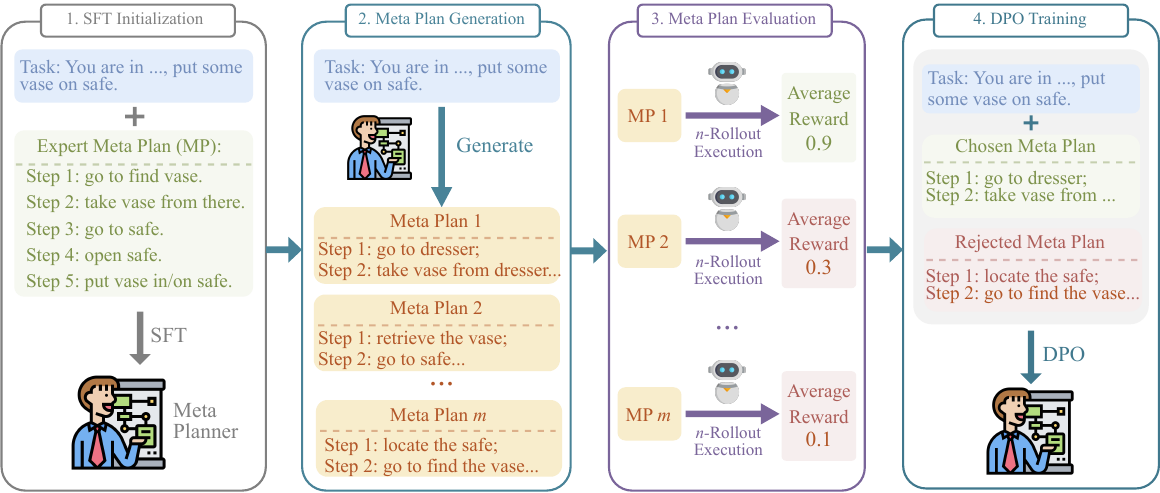}
    \caption{The overall architecture of \method{}. The \planner{} is first supervised fine-tuned on the seed meta plan~(MP) set. Then we optimize the \planner{} through preference learning on contrastive meta plan pairs.}
    \label{fig: main method}
\end{figure*}

\section{Task Formulation}
\paragraph{LLM Agent Planning}
The primary scope of this study is the planning of LLM agents interacting with the environment and receiving feedback for task solution. Following~\citet{song2024trial}, the agent’s task planning trajectory can be represented as \( e = (u, a_1, o_1, \dots, a_n) \), where \( u \in \mathcal{U} \) is the task instruction, \( a \in \mathcal{A} \) the agent actions, and \( o \in \mathcal{O} \) the observation from the environment. At each time step \( t \), the agent performs implicit planning and generates the corresponding action \( a_t \sim \pi_\theta(\cdot | u, a_1, o_1, \dots, o_{t-1})\). The probability of generating the task planning trajectory is given by:

\begin{equation}
\pi_\theta(e | u) = \prod_{t=1}^{n} \pi_\theta(a_t | u, a_1, o_1, \dots, o_{t-1})
\end{equation}
Finally, the final reward \( r(u, e) \in [0, 1] \) representing the task completion rate is calculated.

\paragraph{Meta Plan}
% The \planner{} is a lightweight language model \( \pi_g \) with updatable parameters. Given a task instruction $u$, it generates the task’s meta plan \( p \sim \pi_g(\cdot | u) \). The meta plan is a high-level natural language description outlining the steps required to complete the task. For instance, given the instruction "look at the CD under the desklamp", the generator might produce the following plan: "1. Go to where the CD may be placed. 2. Take the CD from where you found it. 3. Go to where the desklamp is located. 4. Use the desklamp to look at the CD."
% After incorporating the meta plan, the probability of the agent generating trajectory $e$ is formulated as:
% \begin{equation}
% \pi_\theta(e | u, p) = \prod_{t=1}^{n} \pi_\theta(a_t | u, p, a_1, \dots, o_{t-1})\pi_g(p | u)
% \end{equation}
The meta plan serves as high-level, natural guidance to assist in agent planning. It outlines an abstract, general strategy for task completion that is decoupled from specific environmental details, indicating its potential to generalize across various agents.
For instance, given the instruction "look at the CD under the desklamp", the meta plan could be: "1. Go to where the CD may be placed. 2. Take the CD from where you found it. 3. Go to where the desklamp is located. 4. Use the desklamp to look at the CD." 
% This outlines an abstract general task completion strategy that is decoupled from specific environmental details, indicating its potential to generalize across various agents. 
A low-quality meta plan might mislead the agent's planning process. To ensure meta plan quality, \method{} develops a lightweight parameterized \planner{} $\pi_g$ to generate meta plans, which can be further optimized to produce better results. After incorporating the meta plan $p \sim \pi_g( \cdot | u)$, the probability of the agent generating trajectory $e$ is formulated as:
% \begin{equation}
%     p \sim \pi_g(p | u)
% \end{equation}
\begin{equation}
\pi_\theta(e | u, p) = \prod_{t=1}^{n} \pi_\theta(a_t | u, p, a_1, \dots, o_{t-1})
\end{equation}

\section{Method}

The overall framework of our method is illustrated in Figure~\ref{fig: main method}. First, we construct a seed meta plan training set to initialize a basic \planner{}~(\S~\ref{sec: sft inital}). Then, we develop the MC method to assess the quality of the meta plan through exploration~(\S~\ref{sec: metaplan quality evaluation}).  Finally,  we  further enhance the \planner{} via preference-based optimization using contrastive meta plan pairs~(\S~\ref{sec: meta planner optimization}).
\subsection{Supervised Fine-tuning Initialization}
\label{sec: sft inital}
To equip the \planner{} with the foundational capabilities to generate meta plans based on task instructions and the environmental state, we initialize the model using supervised fine-tuning. However, existing agent datasets only provide golden task completion trajectories without corresponding meta plans. Therefore, we first need to construct a training dataset for meta plan generation. To achieve this, we leverage GPT-4o to assist in creating the dataset. We provide the model with the original task instruction $u$ and the corresponding golden trajectory $e$ as the prompt, allowing it to summarize a generalizable plan $p$ from the trajectory. The specific prompt template can be found in Appendix~\ref{appendix: prompt for meta plan collection}. To ensure the quality of the meta plan $p$, we manually review the results generated by GPT-4o and refine any meta plans that are incorrect, overly complex, or non-standard. This quality control process ensures that each meta plan $p$ represents a reusable planning strategy that effectively assists agents in task completion. The detailed process for controlling the quality of the seed meta plan set can be found in Appendix~\ref{appendix: meta plan quality control}. Since the \planner{} needs to generate plans without access to golden trajectories during inference, we remove them from the training data, thus obtaining the initialization dataset for the \planner{}:
\begin{equation}
    \mathcal D_s = \Big\{ (u, p)^{(i)}\Big\}_{i=1}^{|\mathcal D_s|}
\end{equation}
We then fine-tune the model on the auto-regressive loss the get the initialized \planner{} $\pi_g$:
\begin{equation}
    \mathcal L_\mathrm{SFT} = - \mathbb E_{(u, p)\sim\mathcal D_s}[\log\pi_g(p|u)]
\end{equation}

\subsection{Meta Plan Quality Evaluation}
\label{sec: metaplan quality evaluation}
To further enhance the \planner{}, we need to evaluate the quality of its generated meta plans. While prior studies typically rely on reward models trained on human preference annotations~\citep{bai2022training, ouyang2022training, dubey2024llama} or advanced AI~\citep{bai2022constitutional, lee2023rlaif} models to assess model outputs, these approaches have limitations. 
% They not only require additional costs for human labeling or API calls but may also not be directly applicable to LLM agents, as their preferences for meta plans might differ from human preferences. 
They not only incur additional costs for human labeling or API calls, but may also be less applicable to LLM agents, as their preferences for meta plans are not aligned with the agent or task environment.
To circumvent these challenges, we adopt an exploration-based approach to evaluate the quality of meta plans.

Intuitively, a higher-quality meta plan should enable the agent to more easily succeed in the task. 
Therefore, for a give meta plan $p$, we insert it into the prompt of the agent and have the agent attempt to complete the task $N$ times. This results in $N$ task completion trajectories generated by the agent:
\begin{equation}
    \{e^{(i)} | i=1,...,N \} \sim \pi_\theta(e | u, p)
\end{equation}
For each trajectory $e^{(i)}$, the environment returns the task completion rate $r(u, e^{(i)})$. Thus, the quality of the meta plan $p$ is determined by the agent success rate in completing the task based on it, which can be represented as:
\begin{equation}
    Q(p) = \frac{1}{N} \sum_{i=1}^N r(u, e^{(i)})
\end{equation}
In this paper, we use Llama-3.1-8B-Instruct~\citep{dubey2024llama} as the agent to evaluate the quality of the meta plans. This model demonstrates strong instructing-following capabilities and is already effective at completing agent tasks. Moreover, the meta plans evaluated with this model can be generalized to agents based on other models, which we verify in the experiments later.

\subsection{Meta Planner DPO Training}
\label{sec: meta planner optimization}
% After developing the capability to evaluate meta plan quality, we can further optimize the SFT-initialized \planner{} through reinforcement learning. 
After we are able to automatically evaluate the quality of meta plans, we can further optimize the SFT-initialized \planner{} through reinforcement learning.
% We choose DPO~\citep{rafailov2024direct} as our optimization algorithm due to its advantages of training stability and low resource consumption.
We choose DPO~\citep{rafailov2024direct} as our optimization algorithm due to its training stability and low resource consumption. 
The DPO algorithm requires paired preference data to optimize the \planner{}, specifically pairs of high- and low-quality meta plans. 
We construct the DPO preference dataset $\mathcal D_c$ from the task training set, where the SFT-initialized \planner{} generates $M$ meta plans $\{p_i | i=1,...,M \} \sim \pi_g(p|u)$. We then compute scores for each meta plan using the MC method described in Section~\ref{sec: metaplan quality evaluation}. The highest and lowest quality meta plans are selected as the chosen and rejected pairs $p_w$ and $p_l$. If all meta plans are of the same quality, we skip this sample. This forms our preference training dataset:
\begin{equation}
    \mathcal D_c = \Big\{ (u, p_w, p_l)^{(i)}\Big\}_{i=1}^{|\mathcal D_c|}
\end{equation}
Given the preference dataset $D_c$, DPO optimizes the model to increase the likelihood of the chosen meta plan $p_w$ over the rejected one $p_l$. We fine-tune the \planner{} by minimizing the DPO loss:
\begin{equation}
\resizebox{1.0\hsize}{!}{$
\begin{aligned}
\mathcal L_{DPO}(\pi_\theta; \pi_{ref})= - \mathbb E_{(u, p_w, p_l)\sim \mathcal D_c}\bigg[\log\sigma(\beta\log\frac{\pi_\theta(p_w|u)}{\pi_{ref}(p_w|u)} \\
- \beta\log\frac{\pi_\theta(p_l|u)}{\pi_{ref}(p_l|u)} )\bigg],
\end{aligned}
$}
\end{equation}
This equation reflects the goal of maximizing the probability of generating the higher-quality meta plan $p_w$ over the lower-quality meta plan $p_l$ for a given task instruction $u$. By constructing the preference dataset and applying DPO optimization, the \planner{} becomes more effective at generating high-quality meta plans, therefore better guiding the agent planning process.

\begin{table}[t]
    \centering
    \resizebox{\linewidth}{!}{
    \begin{tabular}{l c c c c}
    \toprule
    \textbf{Dataset}   & \textbf{Train} & \textbf{Test Seen} & \textbf{Test Unseen} & \textbf{Action Space}\\
    \midrule
    ScienceWorld & 1483 & 194 & 241 & 19\\
    ALFWorld & 3321 & 140 & 134 & 13\\
    WebShop & 1624 & 200 & - & 8 \\
    \bottomrule
    \end{tabular}
    }
    \caption{Statistics overview of test datasets. “Test Seen” and “Test Unseen” are test set with seen and unseen scenarios
respectively.}
    \label{tab: dataset}
\end{table}

\section{Experiments}

\subsection{Experiment Settings}
\paragraph{Datasets}
\begin{table*}[!htbp]
    \centering
    \resizebox{\linewidth}{!}{
    \begin{tabular}{l c | c c c c c | c}
    \toprule
    \multirow{2}{*}{\textbf{Model}} & \multirow{2}{*}{\textbf{w/o Exp. Guid.}} & \multicolumn{2}{c}{\textbf{ScienceWorld}} & \multicolumn{2}{c}{\textbf{ALFWorld}} & \multicolumn{1}{c |}{\textbf{WebShop}} & \multirow{2}{*}{\textbf{Average}}  \\ \cmidrule(l){3-4} \cmidrule(l){5-6} \cmidrule(l){7-7}
    & & Seen & Unseen & Seen & Unseen & Seen & \\ \midrule
    \multicolumn{7}{c}{\textit{Agents w/o Training}} \\ \midrule
    GPT-4o~\citep{achiam2023gpt} & \ding{55} & 60.0 & 56.0 & 78.6 & 83.6 & 63.5 & 68.3 \\
    GPT-4o-mini~\citep{achiam2023gpt} & \ding{55} & 49.1 & 42.7 & 32.1 & 41.0 & 55.7 & 44.1 \\
    Llama-3.1-8B-Instruct~\citep{dubey2024llama} & \ding{55} & 47.7 & 42.2 & 22.9 & 28.4 & 56.3 & 39.5 \\
    Llama-3.1-8B-Instruct + Reflexion & \ding{55} & 49.6 & 43.1 & 26.5 & 33.1 & 57.4 & 41.9 \\
    Qwen2.5-7B-Instruct~\citep{yang2024qwen2} & \ding{55} & 38.5 & 38.8 & 71.4 & 75.4 & 58.3 & 56.5 \\
    Llama-3.1-70B-Instruct~\citep{dubey2024llama} & \ding{55} & 72.6 & 70.2 & 78.6 & 73.9 & 59.4 & 70.9 \\ \midrule
    \rowcolor[RGB]{233, 255, 233}Llama-3.1-8B-Instruct + \method{} & \ding{51} & 56.5 & 55.5 & 50.0 & 52.2 & 63.2 & 55.5 \\
    \rowcolor[RGB]{233, 255, 233}Llama-3.1-8B-Instruct + Reflexion + \method{} & \ding{51} & 57.5 & 56.4 & 52.0 & 53.4 & 63.9 & 56.6 \\
    \rowcolor[RGB]{233, 255, 233}GPT-4o-mini + \method{} & \ding{51} & 55.7 & 52.8 & 64.3 & 79.9 & 64.0 & 63.3 \\
    \rowcolor[RGB]{233, 255, 233}GPT-4o + \method{} & \ding{51} & 67.3 & 67.8 & \textbf{89.3} & \textbf{93.3} & \textbf{66.3} & 76.8 \\
    \rowcolor[RGB]{233, 255, 233}Llama-3.1-70B-Instruct + \method{} & \ding{51} & \textbf{80.4}  & \textbf{79.5} & 85.7 & 86.6 & 65.1 & \textbf{79.5} \\ \midrule
    \multicolumn{7}{c}{\textit{Agents w/ Training}} \\ \midrule
    Llama-3.1-8B-Instruct + SFT~\citep{zeng2023agenttuning} & \ding{55} & 65.3 & 57.0 & 79.3 & 71.6 & 63.3 & 67.3 \\
    Llama-3.1-8B-Instruct + ETO~\citep{song2024trial} & \ding{55} & 81.3 & 74.1 & 77.1 & 76.4 & 68.4 & 75.5 \\
    Llama-3.1-8B-Instruct + KnowAgent~\citep{zhu2024knowagent} & \ding{51} & 81.7 & 69.6 & 80.0 & 74.9 & 64.8 & 74.2 \\
    Llama-3.1-8B-Instruct + WKM~\citep{DBLP:journals/corr/abs-2405-14205} & \ding{51} & 82.1 & 76.5 & 77.5 & 78.2 & 66.9 & 76.2 \\\midrule
    \rowcolor[RGB]{233, 255, 233}Llama-3.1-8B-Instruct-SFT + \method{} & \ding{51} & 70.2 & 65.9 & 80.7 & \textbf{81.3} & 65.5 & 72.7 \\
    \rowcolor[RGB]{233, 255, 233}Llama-3.1-8B-Instruct-ETO + \method{} & \ding{51} & \textbf{83.4} & \textbf{80.8} & \textbf{85.0} & 79.1 & \textbf{70.2} & \textbf{79.7} \\
    \bottomrule
    \end{tabular}
    }
    \caption{Performance of different methods on two datasets. \method{}-optimized meta plans significantly improve performance across various models or agent frameworks, surpassing other explicit guidance (Exp. Guid.) methods.}
    \label{tab: main results}
\end{table*}
% We conducted experiments on two representative agent datasets: ALFWorld~\citep{shridhar2020alfworld} for embodied household tasks and ScienceWorld~\citep{wang2022scienceworld} for textual science experiment tasks. 
% We conducted experiments on two representative agent datasets: ScienceWorld~\citep{wang2022scienceworld} for textual science experiment tasks and ALFWorld~\citep{shridhar2020alfworld} for embodied household tasks. 
We conducted experiments on three representative agent datasets: ScienceWorld~\citep{wang2022scienceworld} for textual science experiment tasks, ALFWorld~\citep{shridhar2020alfworld} for embodied household tasks, and WebShop for online web navigation tasks~\citep{yao2022webshop}.
% ScienceWorld provides dense final rewards ranging from 0 to 1, whereas ALFWorld offers only binary rewards, indicating whether a task has been completed. For details of the datasets, please refer to Appendix~\ref{appendix: dataset details}.
Both ScienceWorld and WebShop provide dense rewards ranging from 0 to 1, while ALFWorld offers only binary rewards to indicate whether the task is completed. For details of the datasets, please refer to Appendix~\ref{appendix: dataset details}.

The statistical information of our datasets is presented in Table~\ref{tab: dataset}. It is important to note that in addition to the in-distribution test sets, both ALFWorld and ScienceWorld include test sets that include out-of-distribution unseen variations. These additional test sets enable us to evaluate the generalization capabilities of the \planner{}.

\paragraph{Implementation Details}

We use Llama-3.1-8B-Instruct~\citep{dubey2024llama} as the base model to construct the \planner{}. For SFT initialization, we set the batch size to 32, the learning rate to 1e-5 and employ a cosine scheduler with 3 training epochs. For DPO~\citep{rafailov2024direct} training, we configure the \planner{} to generate $M = 5$ meta plans per task with a generation temperature of 0.7. To evaluate meta plan quality, we set the agents to generate $N = 5$ task completion trajectories for each meta plan, also using a temperature of 0.7. We utilize vLLM~\citep{kwon2023efficient} to accelerate the generation process. For DPO training, the batch size is 32, and the learning rate is 1e-5 with a 3\% warm-up phase, and a cosine scheduler is used. The $\beta$ parameter in the DPO loss function is set to 0.1 for the ALFWorld, ScienceWorld and WebShop datasets, with training conducted over 3 epochs. All training procedures are implemented using Llama-Factory~\citep{zheng2024llamafactory} with full parameter fine-tuning. The experiments are conducted on 8 NVIDIA A100 80GB GPUs.

\paragraph{Base Agents}

We evaluate our method on two types of agents, guided by \method{}-optimized meta plans: (1) Agents without training, which deploy the ReAct framework using foundation models without additional training. We test two proprietary models, including GPT-4o and GPT-4o-mini~\citep{achiam2023gpt} as well as several open-source models, including Llama-3.1-8B-Instruct, Llama-3.1-70B-Instruct~\citep{dubey2024llama}, and Qwen2.5-7B-Instruct~\citep{yang2024qwen2}. (2) Agents with training, which enhance agent planning capabilities via parameter updates to foundation models. We examine two agent frameworks: AgentTuning~\citep{zeng2023agenttuning}, which uses Supervised Fine-Tuning from expert trajectories to improve the agent capabilities of the base model, and ETO~\citep{song2024trial}, which learns from failed trajectories and proposes an exploration-based trajectory optimization method to enhance the task-solving process.
We also compare with KnowAgent~\citep{zhu2024knowagent} and WKM~\citep{DBLP:journals/corr/abs-2405-14205}, which also inject explicit guidance into the agent planning process. These two methods require fine-tuning the base models, making them incompatible with other agent frameworks.

\paragraph{Evaluation}

To ensure experimental reproducibility, we set the decoding temperature to 0 for both meta plan generation by the \planner{} and task trajectory generation by the agent. For meta plan generation, we employ a zero-shot prompting approach. When generating task completion trajectories, we include a 1-shot in-context example for each task. The detailed prompts are provided in Appendix~\ref{appendix: prompt for evaluation}. Note that once the meta plans for the test set tasks are generated by the \planner{}, we use them across all agents without further modification. Our primary evaluation metric is the \textbf{Average Reward}, which calculates the mean reward across all test set task instances. We also report the \textbf{Success Rate} in Appendix~\ref{appendix: success rate}. We will release the generated meta plans and parameters of the optimized meta planner upon acceptance.

\subsection{Results}

As shown in Table~\ref{tab: main results}, the incorporation of \method{}-optimized meta plans consistently improves agent performance across all tasks and frameworks, with the average performance increasing by up to 40.5\% for the Llama-3.1-8B-Instruct based agent. Moreover, our \planner{} is compatible with other agent training frameworks. The \method{}-enhanced Llama-3.1-8B-Instruct-ETO achieves an average reward 3.5 higher the the current SOTA explicit guidance method, WKM. 
% These results demonstrate the effectiveness of our method in enhancing agent performance.
These results demonstrate that our general high-level meta plan, optimized through agent feedback, outperforms complex knowledge-based guidance that relies heavily on manual efforts, lacks generalization ability, and offers no quality assurance.
This underscores the effectiveness of our approach in significantly boosting agent performance.
% Furthermore, our method shows strong effectiveness in unseen scenarios. For the unseen part of ScienceWorld and ALFWorld, despite never encountering these tasks, exploration and optimization enable the \planner{} to generate high-quality meta plans.
Furthermore, our method demonstrates strong effectiveness in unseen scenarios. For the unseen parts of ScienceWorld and ALFWorld, despite never having encountered these tasks, the \planner{} is able to generalize to them and generate high-quality meta plans.
This improves the success rate of GPT-4o on the unseen part of ALFWorld by 9.7, achieving a success rate of 93.3. These results highlight that \method{} can further enhance the agent's generalization capabilities, particularly in out-of-distribution scenarios.

\section{Analysis}
\subsection{Ablation Study}
\begin{table}[t]
    \centering
    \resizebox{\linewidth}{!}{
    \begin{tabular}{l | c | c c c}
    \toprule
    \textbf{Base LLM}  &  \textbf{Setting} & \textbf{SciWorld} & \textbf{ALFWorld} & \textbf{WebShop}\\ \midrule
    \multirow{5}{*}{GPT-4o}  & - & 56.0 & 83.6 & 63.5\\
    & SFT & 59.5 & 91.0 & 63.1\\
    & GPT & 60.2 & 91.3 & 64.2\\
    & RFT & 61.8 & 89.6 & 65.5\\
    & \method{} & \textbf{67.8} & \textbf{93.3} & \textbf{66.3}\\ \midrule
    % \multirow{4}{*}{Llama-3.1-8B-Ins} & - & 42.2 & 28.4\\
    % & GPT & 59.5 & 91.0\\
    % & SFT & 45.6 & 43.3\\
    % & RFT & 50.5 & 47.8\\
    % & \method{} & \textbf{55.5} & \textbf{52.2}\\ \midrule
    \multirow{5}{*}{Qwen2.5-7B-Ins}& - & 38.8 & 75.4 & 58.3\\
    & SFT & 37.4 & 73.9 & 59.2\\
    & GPT & 38.3 & 75.6 & 60.8\\
    & RFT & 41.9 & 78.3 & 62.4\\
    & \method{} & \textbf{43.7} & \textbf{82.8} & \textbf{63.6}\\ 
    \bottomrule
    \end{tabular}
    }
    \caption{Ablation study on \planner{} optimization methods. “–” indicates no meta plan. RFT uses reject sampling with only the best sampled meta plan for training. GPT directly prompts GPT-4o as the meta planner.}
    \label{tab: ablation study}
\end{table}
% We conduct ablation experiments on the training methods of the \planner{}. For ScienceWorld and ALFWorld, we evaluate the agent's performance on the unseen test set. As shown in Table~\ref{tab: ablation study}, we find that explicitly injecting meta plans can effectively enhance the agent's planning abilities, thereby improving its overall performance. 

% However, the \planner{} initialized with SFT alone does not consistently lead to performance improvements. For instance, when using SFT-initialized meta plans, the performance of the Qwen2.5-7B model decreases on both evaluation datasets. This may be due to the fact that the meta plans summarized by GPT-4o may not always be suitable for the planning process of all models. In contrast, the RFT and DPO methods, which involve interaction and exploration with the environment, consistently improve the performance of all agents. Additionally, we observe that the performance of RFT trained solely with explored high-quality meta plans still differs from that of \method{} with DPO, highlighting the necessity of optimizing with meta plan quality comparison. This approach explicitly enhances the ability of the \planner{} to distinguish between good and bad meta plans.
We conduct ablation experiments on the training methods of the \planner{}. For ScienceWorld and ALFWorld, we evaluate on the unseen test set. 
% GPT indicates that we use GPT-4o to generate meta-plans for the test set. RFT refers to reject sampling, where only the highest-quality sampled meta-plan is used for training, without leveraging paired meta-plans.
% As shown in Table~\ref{tab: ablation study}, the \planner{} optimized by \method{} leads to greater improvements in agent performance compared to other training methods, suggesting that exploring the environment and learning from comparisons help the \planner{} generate higher-quality meta plans.
As shown in Table~\ref{tab: ablation study}, the \planner{} optimized by \method{} yields greater improvements in agent performance compared to other training methods. It also outperforms directly prompting GPT-4o as the meta planner, which relies solely on its prior knowledge without optimizing the meta plan through environment exploration. This suggests that exploring the environment and learning from comparisons enable the \planner{} to produce higher-quality meta plans.
Additionally, when using SFT-initialized meta plans, the performance of the Qwen2.5-7B-Instruct model decreases on both evaluation datasets, indicating that a low-quality meta plan may mislead the agent planning process.

\subsection{How to Use Meta Plan?}
\begin{table}[t]
    \centering
    \resizebox{\linewidth}{!}{
    \begin{tabular}{l | c | c c}
    \toprule
    \textbf{Base LLM}  &  \textbf{Type} & \textbf{SciWorld} & \textbf{ALFWorld}\\ \midrule
    \multirow{3}{*}{GPT-4o}  & Inst. & \textbf{67.8} & \textbf{93.3} \\
    & Thou. & 65.3 & 85.1 \\
    & Obs. & 67.6 & 91.8 \\ \midrule
    \multirow{3}{*}{Llama-3.1-8B} & Inst. & \textbf{55.5} & \textbf{52.2}\\
    & Thou. & 38.0 & 34.3\\
    & Obs. & 53.3 & 50.8\\ \midrule
    \multirow{3}{*}{Llama-3.1-8B-SFT}& Inst. & \textbf{65.9} & \textbf{81.3} \\
    & Thou. & 47.9 & 25.4 \\
    & Obs. & 60.6 & 67.2 \\ 
    \bottomrule
    \end{tabular}
    }
    \caption{The impact of different meta plan insertion positions on agent performance.}
    \label{tab: meta plan positon}
\end{table}
\begin{figure}
    \centering
    \includegraphics[width=\linewidth]{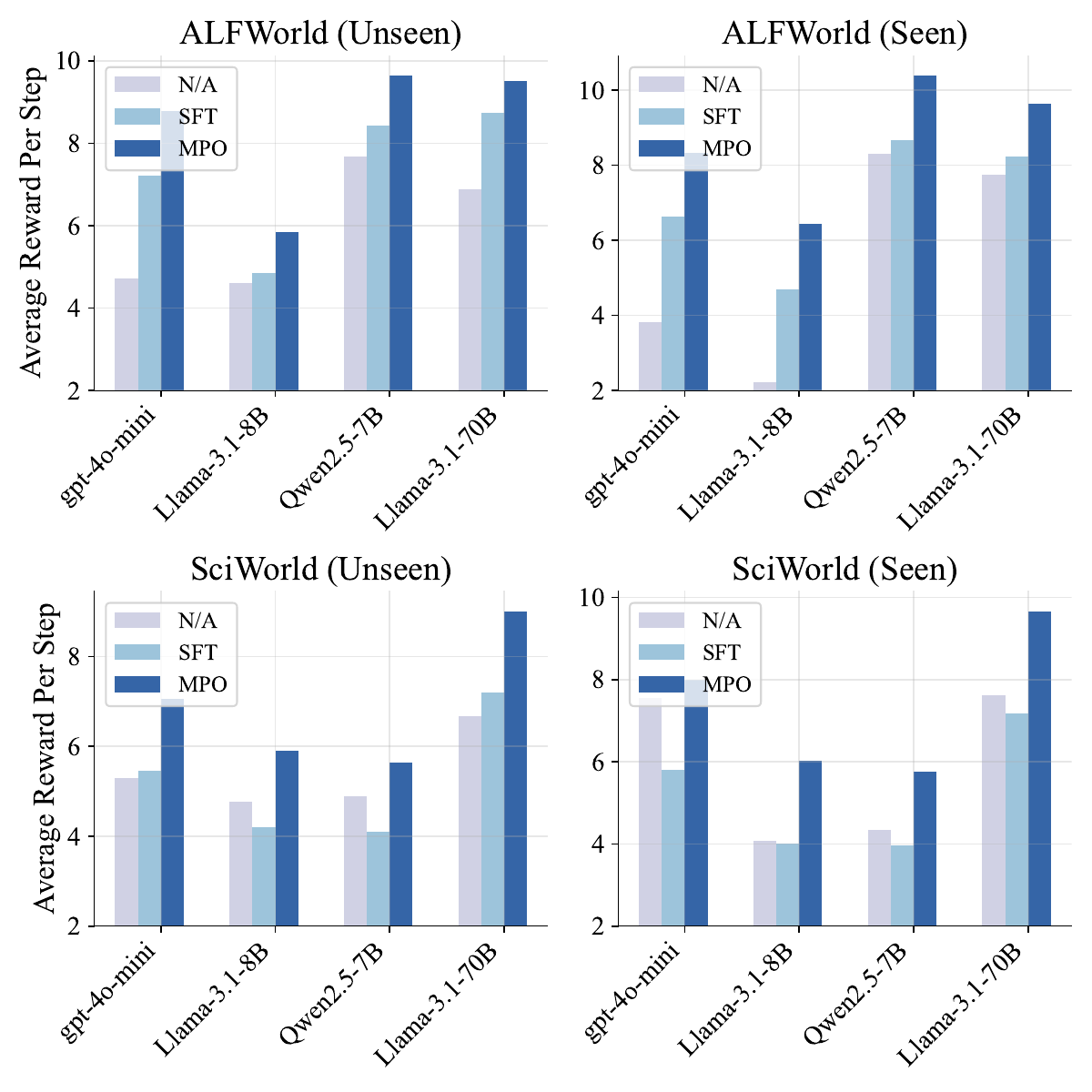}
    \caption{The average reward per step.}
    \label{fig: step reward}
\end{figure}

In our main experiments, the meta plan is incorporated into the task instructions to guide the agent planning process. Here, we investigate the impact of different insertion positions on agent performance: in the task instruction, in the agent's thought process and in the environment observation. As shown in Table~\ref{tab: meta plan positon}, we find that insertion into the task instruction consistently yields the best performance across all agents and tasks, while insertion into the thought process leads to the worst performance. This suggests that disrupting the agent's normal reasoning process negatively affects planning accuracy. Additionally, we observe that inserting the meta plan at other positions causes greater performance drops in agent frameworks with training, likely because the training data does not involve meta plans. In contrast, insertion into the instruction causes minimal disruption to the original task completion process. These results suggest that inserting the meta plan in the task instruction ensures optimal performance.

\subsection{Efficiency Analysis}

\begin{figure}
    \centering
    \includegraphics[width=0.98\linewidth]{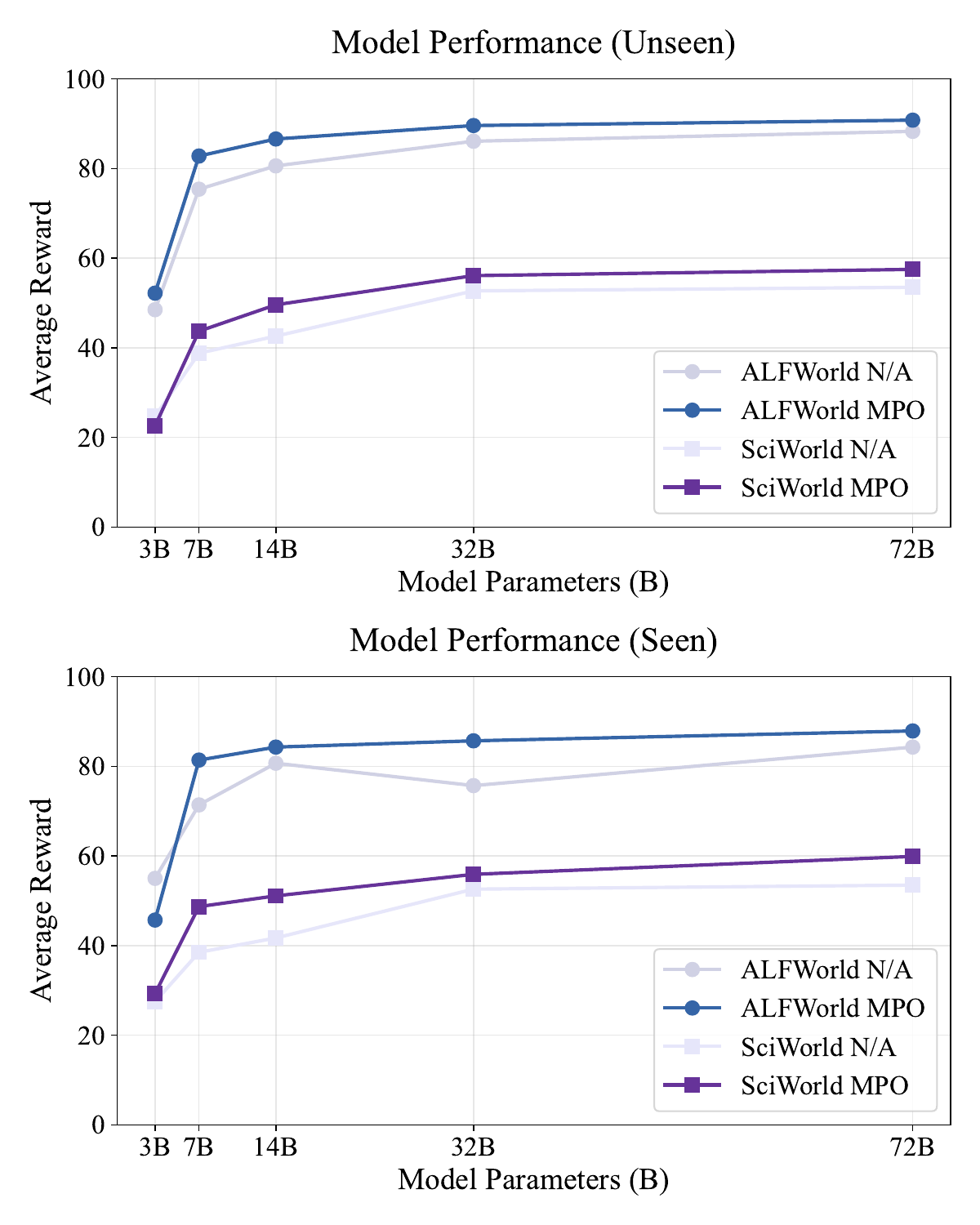}
    \caption{The effectiveness of \method{} across agents with different parameter sizes.}
    \label{fig: scaling effect}
\end{figure}

% Another advantage of incorporating high-quality meta plans is that it prevents agents from engaging in unnecessary exploration in the environment, thereby improving the their efficiency in completing tasks. We evaluate the action efficiency of the agent using the average reward per step. We calculate it for each tasks as the ratio of the final reward to the number of steps required to complete the tasks, and then average these values across the entire test set. Figure~\ref{fig: step reward} illustrates the significant improvements in average step rewards achieved by our \method{} method compared to both the no-meta-plan and SFT-initialized \planner{}s. It is also evident that for the unseen test tasks, \method{} can provide an even greater increase in average reward per step, indicating its strong generalization to out-of-distribution tasks. These results underscores the superior performance of \method{}, confirming its effectiveness in enhancing agent action efficiency.

Another advantage of incorporating high-quality meta plans is that it prevents agents from unnecessary exploration, thus improving their task completion efficiency. Following~\citet{xiong2024watch}, we evaluate action efficiency using the average reward per step, calculated for each task as the ratio of the final reward to the number of steps required to complete the task, and then averaging these values across the entire test set. Figure~\ref{fig: step reward} shows the significant improvements in average step rewards achieved by our \method{} compared to both the no-meta-plan~(N/A) and SFT-initialized meta plans. It is also clear that for the unseen test tasks, \method{} leads to an even greater increase in average reward per step, demonstrating its strong generalization to out-of-distribution tasks. These results underscore the superior performance of \method{}, confirming its effectiveness in enhancing agent action efficiency.

\subsection{Effect on Agents with Scaling Parameters}

% To further validate the effectiveness of our method, we conduct experiments on models with different parameter sizes. We choose Qwen2.5-Instruct family as the test models, selecting a range of parameter sizes from small to large: 3B, 7B, 14B, 32B, and 72B, and evaluate their performance on the ScienceWorld and ALFWorld seen and unseen parts. The effectiveness of \method{} across agents with different parameter sizes is shown in Figure~\ref{fig: scaling effect}. We observe that as the parameter size of the act agents increases, the performance improvement brought by \method{} follows a trend of initially increasing and then decreasing. This may be because the 72B model already has a strong planning ability, making the improvement from \method{} relatively limited. Additionally, for the 3B model, due to its limited instruction-following ability, the model struggles to effectively follow the met plan. As a result, inserting the meta plan into the prompt actually leads to a decrease in performance. These results indicate that \method{} can enhance the performance of agents across a wide range of parameter sizes, with the most significant improvements observed in agents with medium-sized parameters.
To further validate the effectiveness of our method, we conduct experiments on models of varying parameter sizes. We choose the Qwen2.5-Instruct family as test models, selecting a range of parameter sizes from small to large: 3B, 7B, 14B, 32B, and 72B, and evaluate their performance on both the seen and unseen parts of ScienceWorld and ALFWorld. 

% The effectiveness of \method{} across agents with different parameter sizes is shown in Figure~\ref{fig: scaling effect}. We observe that as the parameter size of the agents increases, the performance improvement from \method{} initially increases and then decreases. This may be because the 72B model already has strong planning capabilities, making the improvement from \method{} relatively limited. Additionally, for the 3B model, due to its limited instruction-following ability, the model struggles to effectively utilize the meta plan. As a result, inserting the meta plan into the prompt actually leads to a performance decrease. 
% These results suggest that \method{} can enhance agent performance across a wide range of parameter sizes, with the most significant improvements observed in agents with medium-sized parameters. Moreover, as a lightweight model, the meta planner has the potential to enhance more powerful agents portably without the need for retraining the agents.
As shown in Figure~\ref{fig: scaling effect}, \method{} can enhance agent performance across a wide range of parameter sizes, with the most significant improvements observed in agents with medium-sized parameters. Moreover, as a lightweight model, the meta planner has the potential to enhance more powerful agents portably without requiring their retraining.
\begin{figure}
    \centering
    \includegraphics[width=\linewidth]{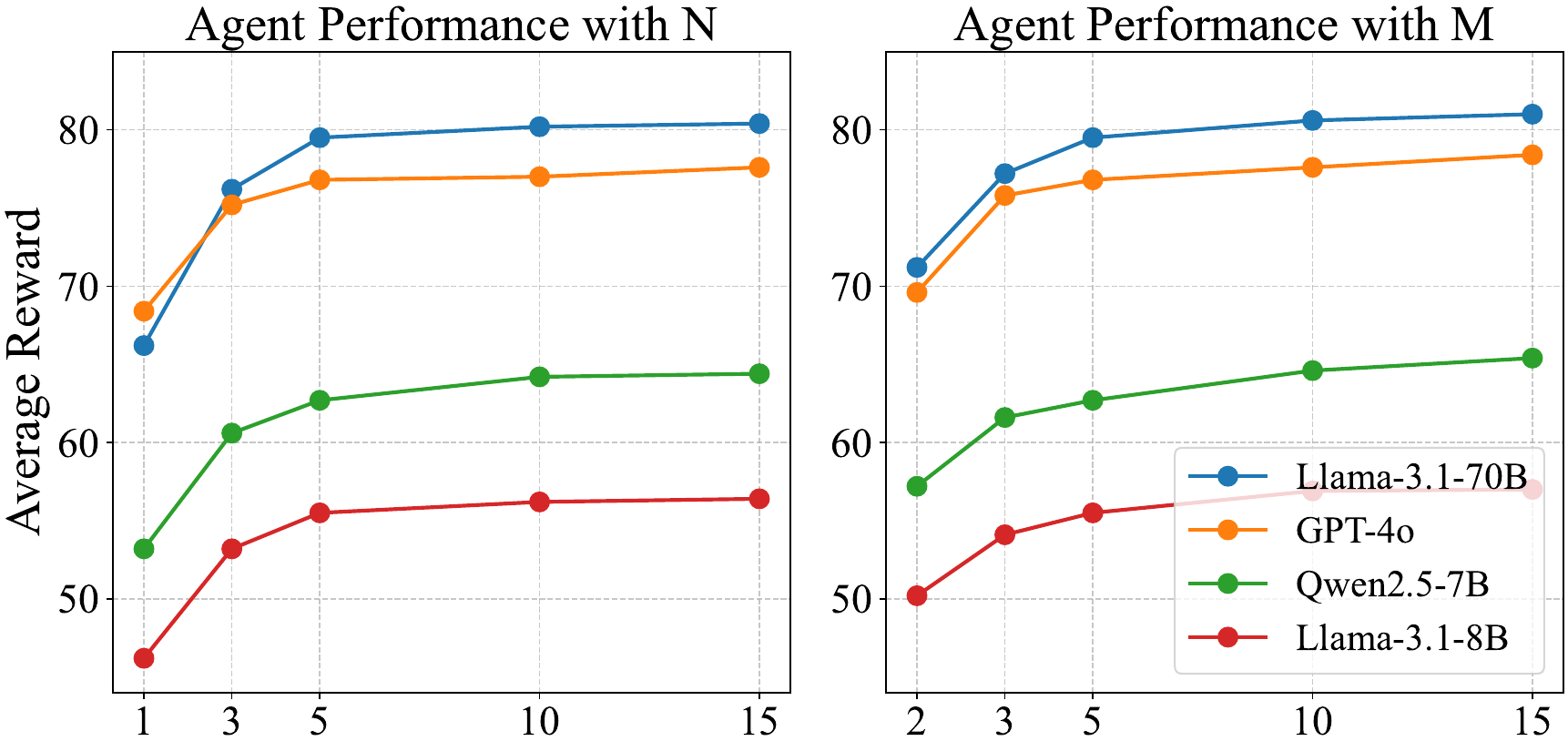}
    \caption{Impact of sampling count \(N\) and \(M\) in contrastive meta plan construction on agent performance.}
    \label{fig: sample count}
\end{figure}
\begin{figure*}
    \centering
    \includegraphics[width=\linewidth]{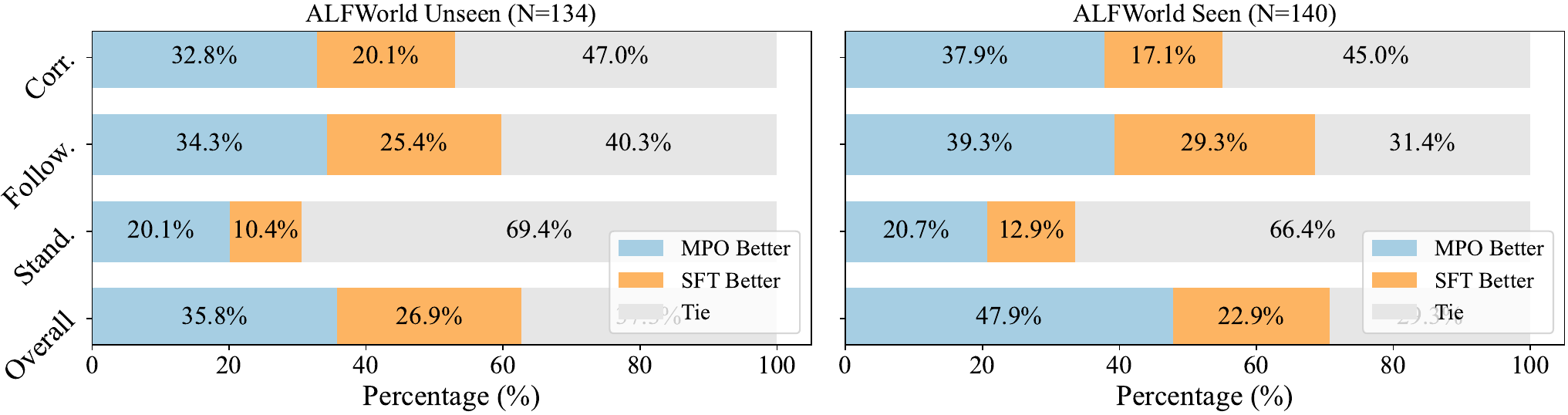}
    \caption{The comparison of SFT-initialized and MPO-optimized meta plans on ALFWorld.}
    \label{fig: meta plate compare}
\end{figure*}

\subsection{Effect of Sampling Count on Performance}
% We analyze the impact of different sampling counts—$M$ for meta plan sampling and $N$ for task completion trajectory sampling—on the performance of the trained meta planner. We use the average performance of the agent enhanced by the optimized meta plans as the evaluation metric, and assess the effect of each parameter by fixing one while varying the other. 
% As shown in Figure~\ref{fig: sample count}, lower sampling counts significantly degrade the quality of the meta plans, with $N$ having a larger impact. This is likely because insufficient task trajectory samples reduce the accuracy of meta plan quality estimation. Additionally, increasing $N$ and $M$ yields diminishing returns, with agent performance quickly converging. 
% Despite the efficiency concerns of sampling, we find a good balance between sample size and meta planner performance by setting $N = M = 5$ in the main experiments, achieving strong results with minimal overhead.
We analyze how the sampling counts, $M$ for meta plan generation and $N$ for task trajectory sampling affect the trained meta planner's performance. We use the average score of the agent enhanced by the optimized meta plans as the evaluation metric, varying one parameter while fixing the other.
As shown in Figure~\ref{fig: sample count}, lower sampling counts significantly reduce meta plan quality, with $N$ having a stronger effect. This is likely because insufficient task trajectories reduce the accuracy of meta plan quality estimation. Moreover, the performance gains from increasing $N$ and $M$ quickly saturate.
Considering sampling efficiency, we set $N$ = $M$ = 5 in the main experiments, striking a balance between sampling cost and meta planner performance gains.

\subsection{What Makes a Good Meta Plan?}

We further investigate why the meta plans optimized through exploration in \method{} outperform those obtained soly through SFT initialization. We evaluate the meta plans from three perspectives: correctness, followability, and standardization, using GPT-4o for automated assessment. The evaluation details and prompts can be found in Appendix~\ref{appendix: prompt for GPT automated assessment}. As shown in Figure~\ref{fig: meta plate compare}, \method{}-optimized meta plans consistently outperform SFT-initialized ones across all three dimensions. The advantages in correctness and followability make it easier for the agent to effectively plan and execute tasks, leading to higher task completion rates. Please refer to Appendix~\ref{appendix: case study} for a more detailed case study.

\section{Related Work}
\paragraph{LLM as Agents}
% Agent is an automated system deployed in an external environment, capable of interacting with the environment and acting upon it~\citep{franklin1997autonomous}. The development of autonomous agents is considered a promising path toward achieving Artificial General Intelligence~(AGI)~\citep{wang2024survey}. With the advancements in reasoning and instruction-following capabilities of large language models, researchers have begun using prompting methods~\citep{wei2022chain, yao2022react} or more complex prompting strategies~\citep{song2023restgpt} to enable agents to leverage tools, solve problems, write codes, and complete real-world tasks~\citep{patil2023gorilla, gur2023real, yang2024swe}. To enhance the capabilities of open-source models as agents, recent efforts~\citep{chen2023fireact, zeng2023agenttuning} have leveraged pre-constructed expert trajectories as multi-turn conversation data to fine-tune LLMs, a technique known as behavioral cloning~\citep{pomerleau1991efficient}. To further improve agent capabilities, some works~\citep{song2024trial, xiong2024watch} have started enabling agents to explore the environment autonomously and use reinforcement learning to utilize contrastive information from failed trajectories. However, these methods require updating all parameters of the LLM, and retraining is necessary each time a new model is deployed, which incurs significant overhead. Unlike this, our work proposes a plug-and-play approach that directly enhances the agent's capabilities without introducing additional overhead.
With advancements in reasoning and instruction-following capabilities of large language models~\citep{wei2022emergent}, researchers have begun using prompting methods~\citep{wei2022chain, song2023restgpt} or more complex strategies~\citep{koh2024tree} to build agents capable of leveraging tools~\citep{qin2023toolllm}, solving problems, writing code~\citep{qian2023communicative}, and completing real-world tasks~\citep{patil2023gorilla, gur2023real, yang2024swe}. 
% To enhance the capabilities of open-source models as agents, recent efforts~\citep{chen2023fireact, zeng2023agenttuning, song2024trial, xiong2024watch} have started using expert trajectories for supervised fine-tuning of LLMs or enabling them to explore environments autonomously for learning. 
To enhance open-source models as agents, some works~\citep{zeng2023agenttuning, song2024agentbank} use expert trajectories for supervised fine-tuning LLMs, while others~\citep{song2024trial, xiong2024watch, zhao2024epo} enable agents to explore the environment autonomously and leverage reinforcement learning to learn from failed experiences.
However, these methods require retraining each time a new agent is deployed, leading to significant computational overhead.

\paragraph{Planning in LLM Agents}

Planning~\citep{huang2024understanding} is essential for intelligent agents to complete real-world tasks, involving the decomposition of complex instructions into sub-tasks and acting on them sequentially. Previous works~\citep{yao2022react, shinn2024reflexion} primarily focus on implicit planning, where planning occurs through interleaved reasoning and action generation. To address the challenges of myopic reasoning and planning hallucination in implicit planning~\citep{zhu2024knowagent}, some approaches~\citep{guan2024amor, li2024formal, zhao2024expel, zhu2024knowagent} have explored using explicit knowledge to guide task execution. 
However, these methods often require manually designed prompt templates or task procedures, limiting their transferability across environments.
Some works~\citep{logeswaran2022few, ye2023proagent, fu2024autoguide} use language models to automate task knowledge synthesis or subgoal planning, but the generated knowledge cannot be further refined through exploration and environmental feedback, leading to suboptimal performance. 
% In contrast, our \method{} not only explicitly models and automates the planning process but also allows for the optimization of meta plan generation based on feedback from the agent's task execution. 
In contrast, our \method{} introduces an automatically generated meta plan that provides high-level, abstract guidance to assist in agent planning, while allowing further quality enhancement based on feedback from the task completion process.

% A comparison of \method{} with several alternatives in Table~\ref{tab: comparation} highlights the advantages of our method in enhancing LLM agents planning capabilities.

\section{Conclusion}
% In this paper, we present \method{}, a novel framework for enhancing the planning capabilities of LLM-based agents. 
% \method{} integrates abstract, high-level guidance through meta plans, providing  a plug-and-play solution to efficiently improve agent performance. 
In this paper, we introduce \method{}, a novel framework for enhancing the performance of LLM-based agents.
\method{} incorporates abstract, high-level guidance via meta plans, offering an innovative solution to explicitly improve the agent's planning capabilities.
By utilizing feedback from the agent's task execution, \method{} enables continuous enhancement of the meta plan quality. Extensive experiments on three benchmarks demonstrate that our framework consistently outperforms existing baselines and is applicable to agents  across a wide range of parameter sizes. These findings highlight the potential of our approach to advance agent planning capabilities, paving the way for future developments in artificial general intelligence.

% \clearpage
\section*{Acknowledgement}
We thank the anonymous reviewers for their helpful comments on this paper. 
This work was partially supported by National Natural Science Foundation of China project (No. 62476010).

\section*{Limitations}
Despite achieving superior performance compared to other baselines, it is important to acknowledge several limitations of this work. 
1) Our approach uses Llama-3.1-8B-Instruct as the base model to construct the \planner{}. However, it is worth exploring the potential differences when utilizing other base models or models with varying parameter sizes for the meta planner. Future work could investigate the use of more lightweight models, such as those with as few as 1B parameters, to enhance computational efficiency. 
2) Our method only focuses on constructing a separate meta planner for each individual task. However, building a meta planner that incorporates data from multiple tasks may allow it to learn from diverse knowledge sources, resulting in higher-quality meta plans. Future research could develop a unified meta planner that is applicable to various tasks. 
3) In the \planner{} DPO training, we employ simple sampling and Monte Carlo methods to construct contrastive meta plan pairs. Future work could explore the application of MCTS methods to improve the efficiency of the sampling process.

\section*{Ethics Statement}
This work fully complies with the ACL Ethics Policy.
We declare that there are no ethical issues in this paper, to the best of our knowledge.

\bibliography{custom}

\clearpage
\appendix

\section{Dataset Details}
\label{appendix: dataset details}
\paragraph{ScienceWorld}

\begin{table*}[!htbp]
    \centering
    \resizebox{\linewidth}{!}{
    \begin{tabular}{l c | c c c c | c}
    \toprule
    \multirow{2}{*}{\textbf{Model}} & \multirow{2}{*}{\textbf{w/o Meta Plan}} & \multicolumn{2}{c}{\textbf{ScienceWorld}} & \multicolumn{2}{c|}{\textbf{ALFWorld}} & \multirow{2}{*}{\textbf{Average}}  \\ \cmidrule(l){3-4} \cmidrule(l){5-6}
    & & Seen & Unseen & Seen & Unseen & \\ \midrule
    \multicolumn{7}{c}{\textit{Agents w/o Training}} \\ \midrule
    \multirow{2}{*}{GPT-4o~\citep{achiam2023gpt}} & \ding{55} & 60.0 & 56.0 & 78.6 & 83.6 & 69.6 \\
    & \ding{51} & 67.3 & 67.8 & 89.3 & 93.3 & 79.4 \\ \midrule
    \multirow{2}{*}{GPT-4o-mini~\citep{achiam2023gpt}} & \ding{55} & 49.1 & 42.7 & 32.1 & 41.0 & 41.2 \\
    & \ding{51} & 55.7 & 52.8 & 64.3 & 79.9 & 63.2 \\ \midrule
    \multirow{2}{*}{Llama-3.1-8B-Instruct~\citep{dubey2024llama}} & \ding{55} & 47.7 & 42.2 & 22.9 & 28.4 & 35.3 \\
    & \ding{51} & 56.5 & 55.5 & 50.0 & 52.2 & 53.6 \\ \midrule
    \multirow{2}{*}{Qwen2.5-7B-Instruct~\citep{yang2024qwen2}} & \ding{55} & 38.5 & 38.8 & 71.4 & 75.4 & 56.0 \\
    & \ding{51} & 41.7 & 43.7 & 81.4 & 82.8 & 62.4 \\ \midrule
    \multirow{2}{*}{Llama-3.1-70B-Instruct~\citep{dubey2024llama}} & \ding{55} & 72.6 & 70.2 & 78.6 & 73.9 & 73.8 \\
    & \ding{51} & 80.4 & 79.5 & 85.7 & 86.6 & 83.1 \\ \midrule
    \multicolumn{7}{c}{\textit{Agents w/ Training}} \\ \midrule
    \multirow{2}{*}{Llama-3.1-8B-Instruct + SFT~\citep{zeng2023agenttuning}} & \ding{55} & 65.3 & 57.0 & 79.3 & 71.6 & 68.3 \\
    & \ding{51} & 70.2 & 65.9 & 80.7 & 81.3 & 74.5 \\ \midrule
    \multirow{2}{*}{Llama-3.1-8B-Instruct + ETO~\citep{song2024trial}} & \ding{55} & 81.3 & 74.1 & 77.1 & 76.4 & 77.2 \\
    & \ding{51} & 83.4 & 80.8 & 85.0 & 79.1 & 82.1 \\
    \bottomrule
    \end{tabular}
    }
    \caption{The average reward comparison of different agents after incorporating \method{}-optimized meta plans on two datasets.}
    \label{tab: average reward}
\end{table*}

\begin{table*}[!htbp]
    \centering
    \resizebox{\linewidth}{!}{
    \begin{tabular}{l c | c c c c | c}
    \toprule
    \multirow{2}{*}{\textbf{Model}} & \multirow{2}{*}{\textbf{w/o Meta Plan}} & \multicolumn{2}{c}{\textbf{ScienceWorld}} & \multicolumn{2}{c|}{\textbf{ALFWorld}} & \multirow{2}{*}{\textbf{Average}}  \\ \cmidrule(l){3-4} \cmidrule(l){5-6}
    & & Seen & Unseen & Seen & Unseen & \\ \midrule
    \multicolumn{7}{c}{\textit{Agents w/o Training}} \\ \midrule
    \multirow{2}{*}{GPT-4o~\citep{achiam2023gpt}} & \ding{55} & 59.8 & 57.8 & 78.6 & 83.6 & 70.0 \\
    & \ding{51} & 61.3 & 65.9 & 89.3 & 93.3 & 77.5 \\ \midrule
    \multirow{2}{*}{GPT-4o-mini~\citep{achiam2023gpt}} & \ding{55} & 38.7 & 28.9 & 32.1 & 41.0 & 35.2 \\
    & \ding{51} & 41.2 & 41.2 & 64.3 & 79.9 & 56.7 \\ \midrule
    \multirow{2}{*}{Llama-3.1-8B-Instruct~\citep{dubey2024llama}} & \ding{55} & 25.8 & 25.6 & 22.9 & 28.4 & 25.7 \\
    & \ding{51} & 47.9 & 53.6 & 50.0 & 52.2 & 50.9 \\ \midrule
    \multirow{2}{*}{Qwen2.5-7B-Instruct~\citep{yang2024qwen2}} & \ding{55} & 22.7 & 30.8 & 71.4 & 75.4 & 50.1 \\
    & \ding{51} & 32.0 & 33.2 & 81.4 & 82.8 & 57.4 \\ \midrule
    \multirow{2}{*}{Llama-3.1-70B-Instruct~\citep{dubey2024llama}} & \ding{55} & 67.5 & 64.9 & 78.6 & 73.9 & 71.2 \\
    & \ding{51} & 71.7 & 69.7 & 85.7 & 86.6 & 78.4 \\ \midrule
    \multicolumn{7}{c}{\textit{Agents w/ Training}} \\ \midrule
    \multirow{2}{*}{Llama-3.1-8B-Instruct + SFT~\citep{zeng2023agenttuning}} & \ding{55} & 59.3 & 64.9 & 79.3 & 71.6 & 68.8 \\
    & \ding{51} & 68.6 & 72.0 & 80.7 & 81.3 & 75.7 \\ \midrule
    \multirow{2}{*}{Llama-3.1-8B-Instruct + ETO~\citep{song2024trial}} & \ding{55} & 75.8 & 77.7 & 77.1 & 76.4 & 76.8 \\
    & \ding{51} & 80.9 & 78.7 & 85.0 & 79.1 & 80.9 \\
    \bottomrule
    \end{tabular}
    }
    \caption{The success rate comparison of different agents after incorporating \method{}-optimized meta plans on two datasets. For ALFWorld, the success rate is equivalent to the average final reward.}
    \label{tab: success rate}
\end{table*}
ScienceWorld~\citep{wang2022scienceworld} is a text-based virtual environment that provides a testing platform for AI research, specifically designed to evaluate and improve AI systems' scientific reasoning abilities. Researchers can use this platform to assess the performance of AI agents in open, complex environments. ScienceWorld simulates tasks from standard elementary school science curricula, covering areas such as state changes of matter, measurement, electricity, life sciences, plant growth, chemical reactions, and genetics. Agents are deployed in an embodied interactive environment to understand and apply complex scientific concepts. Tasks in ScienceWorld involve several subgoals, and the overall final reward is calculated based on the completion of these subgoals.

The original test set of ScienceWorld includes unseen task variations. For example, in the training set, a task may involve boiling water, while in the test set, the task may be boiling lead. Following~\citet{song2024trial}, we use the original test set to evaluate the generalization ability of our meta planner in unseen scenarios, and the original validation set serves as our test set for seen scenarios.

\paragraph{ALFWorld}
ALFWorld~\citep{shridhar2020alfworld} are household tasks that require agents to explore rooms and use commonsense reasoning to perform tasks, such as "put a pencil on the desk". The environment provides the outcome on whether the agent successfully completes the task within given steps. The original ALFWorld dataset comprises both seen and unseen evaluation sets. The seen set is designed to assess in-distribution generalization, whereas the unseen set with new tasks measures out-of-distribution generalization of the agents.

\paragraph{WebShop}
WebShop~\citep{yao2022webshop} is a simulated e-commerce website environment containing 1.18 million real-world products. In this environment, agents are required to navigate through various types of webpages and perform diverse actions to find, customize, and purchase items based on natural language instructions. Once the agent clicks the "buy" option, the environment provides a final reward, which is calculated based on the matching heuristics of the product’s attributes and price.

\section{Success Rate}
\label{appendix: success rate}
We report the success rate of our experiments in Table~\ref{tab: success rate}. Note that the definition of success rate differs between the two tasks. For ScienceWorld, the original paper does not provide a specific definition for success rate. However, based on the official environment, a trajectory is considered successful if the agent reaches a predefined latent state, even if the reward is not exactly 1.0. For ALFWorld, since it only provides binary final rewards, the success rate is equivalent to the average final reward. After inserting the \method{}-optimized meta plan, all agents show consistent and significant success rate improvements across both tasks.

\section{Seed Meta Plans Quality Control}
\label{appendix: meta plan quality control}
A high-quality seed meta plan training set is crucial for initializing a more effective meta planner. As such, we carefully control the quality of the meta plans generated by GPT-4o. We have identified several key issues with the meta plans it produces: (1) they often include excessively detailed steps or environmental information, which makes them difficult to generalize and optimize; (2) they sometimes feature manipulation types that are not applicable to the environment; (3) they fail to adhere to the predefined meta plan format. To address the first two issues, we adjust the temperature during GPT-4o's generation and re-summarize the meta plan. 
For the third issue, we additionally prompt GPT-4o to extract correctly formatted meta plans from the response.
Although manual verification is required to ensure quality, the human effort involved in this process is negligible compared to the manual construction of knowledge in \citet{zhu2024knowagent}.

\section{Case Study}
\label{appendix: case study}
Here we provide a detailed comparison of agent trajectories on the same task within ALFWorld, after inserting meta plans optimized by two different methods: SFT and \method{}. This comparison demonstrates how \method{} provides higher-quality plan guidance. The case is shown in Figure~\ref{fig:alfworld_case}. The agent used in this case study is Llama-3.1-8B-Instruct.

In the ALFWorld scenario, the meta plan generated by the SFT-initialized meta planner mistakenly includes the instruction "go to sidetable", which misleads the agent into repeatedly executing the erroneous plan "I can try to go to sidetable first," resulting in plan hallucination. In contrast, the \method{}-optimized meta planner generates a higher-quality meta plan: "go to where the first pillow may be located." This plan outlines an abstract, general task completion strategy, decoupled from specific environmental details, and correctly guides the agent in planning to locate the pillow in the environment with "I can check one by one, starting from armchair 1."

\section{Prompts Used in Our Work}
\subsection{Prompt for Seed Meta Plans Collection}
\label{appendix: prompt for meta plan collection}
We show the prompt for GPT-4o to generate the seed meta plan dataset based on the task instructions. We provide the task instruction, environmental information, and the current task completion trajectory, then prompt GPT-4o to extract a meta plan that includes environmental priors and can guide the task completion process. The prompt is shown in Figure~\ref{fig: sciworld meta plan prompt}, Figure~\ref{fig: alfworld meta plan prompt} and Figure~\ref{fig: webshop meta plan prompt}.

\subsection{Prompt for Evaluation}
\label{appendix: prompt for evaluation}
We show the instruction prompts for ScienceWorld, ALFWorld and WebShop in Figure~\ref{fig: scienceworld prompt}, ~\ref{fig: alfworld prompt} and~\ref{fig: webshop prompt}, respectively.

\subsection{Prompt for GPT Automated Assessment}
\label{appendix: prompt for GPT automated assessment}
We show the prompt in Figure~\ref{fig: prompt for gpt automated assessment} that enables GPT-4o to automatically evaluate the quality of the MPO-optimized meta plan from three aspects: correctness, followability, and standardization. Correctness assesses whether the plan accurately fulfills the task requirements, followability evaluates whether the plan is clear, easy to understand, and whether the steps are reasonable, while standardization checks if the meta plan follows a consistent and standardized format. For each dimension, GPT-4o is asked to first identify which set of plans is better and provide the reasoning procedure. Finally, an overall assessment is given.

\onecolumn

\begin{tcolorbox}[breakable,title=Prompt for ScienceWorld Meta Plan Collection]
Please generate a step-by-step meta plan for a scientific task:\\
<task>\\
You are a helpful assistant to do some scientific experiment in an environment.\\
In the environment, there are several rooms: kitchen, foundry, workshop, bathroom, outside, living room, bedroom, greenhouse, art studio, hallway.\\
\{task\}\\
</task>\\

You should explore the environment and find the items you need to complete the experiment.
You can teleport to any room in one step.\\
All containers in the environment have already been opened, you can directly get items from the containers.\\

The available actions are:\\
    \hspace*{1em} open OBJ: open a container\\
    \hspace*{1em} close OBJ: close a container\\
    \hspace*{1em} activate OBJ: activate a device\\
    \hspace*{1em} deactivate OBJ: deactivate a device\\
    \hspace*{1em} connect OBJ to OBJ: connect electrical components\\
    \hspace*{1em} disconnect OBJ: disconnect electrical components\\
    \hspace*{1em} use OBJ [on OBJ]: use a device/item\\
    \hspace*{1em} look around: describe the current room\\
    \hspace*{1em} examine OBJ: describe an object in detail\\
    \hspace*{1em} look at OBJ: describe a container's contents\\
    \hspace*{1em} read OBJ: read a note or book\\
    \hspace*{1em} move OBJ to OBJ: move an object to a container\\
    \hspace*{1em} pick up OBJ: move an object to the inventory\\
    \hspace*{1em} pour OBJ into OBJ: pour a liquid into a container\\
    \hspace*{1em} mix OBJ: chemically mix a container\\
    \hspace*{1em} teleport to LOC: teleport to a specific room\\
    \hspace*{1em} focus on OBJ: signal intent on a task object\\
    \hspace*{1em} wait: task no action for 10 steps\\
    \hspace*{1em} wait1: task no action for a step\\

Below is the standard and detailed procedure for solving this task:\\
<conversation>\\
\{conversation\}\\
</conversation>\\

You need to conclude abstract steps as a meta plan, which can be used to solve similar tasks in the future.\\
The meta plan should be a commonly-reused routine of the tasks.\\
The generated meta plan should be written in the following format:\\
<meta\_plan>\\
Step 1: ...\\
Step 2: ...\\
...\\
</meta\_plan>
\end{tcolorbox}
\begin{figure*}[ht]
    \centering
    \vspace{-8pt}
    \caption{
    Prompt for ScienceWorld Meta Plan Collection.
    }
    \label{fig: sciworld meta plan prompt}
\end{figure*}

\begin{tcolorbox}[breakable,title=Prompt for ALFWorld Meta Plan Collection]
Please generate a step-by-step meta plan for a house holding task:\\
<task>\\
\{task\}\\
</task>\\

The action list you can take:\\
\hspace*{1em} 1. go to {{recep}}\\
\hspace*{1em} 2. task {{obj}} from {{recep}}\\
\hspace*{1em} 3. put {{obj}} in/on {{recep}}\\
\hspace*{1em} 4. open {{recep}}\\
\hspace*{1em} 5. close {{recep}}\\
\hspace*{1em} 6. toggle {{obj}} {{recep}}\\
\hspace*{1em} 7. clean {{obj}} with {{recep}}\\
\hspace*{1em} 8. heat {{obj}} with {{recep}}\\
\hspace*{1em} 9. cool {{obj}} with {{recep}}\\
where {{obj}} and {{recep}} correspond to objects and receptacles.\\

Below is the standard and detailed procedure for solving this task:\\
<conversation>\\
\{conversation\}\\
</conversation>\\

The generated meta plan should be written in the following format:\\
<meta\_plan>\\
Step 1: ...\\
Step 2: ...\\
...\\
</meta\_plan>
\end{tcolorbox}
\begin{figure*}[ht]
    \centering
    \vspace{-8pt}
    \caption{
    Prompt for ALFWorld Meta Plan Collection.
    }
    \label{fig: alfworld meta plan prompt}
\end{figure*}

\begin{tcolorbox}[breakable,title=Prompt for WebShop Meta Plan Collection]
Please generate a step-by-step meta plan for a webshopping task:\\
You are web shopping. I will give you instructions about what to do. You have to follow the instructions.\\
<task>\\
\{task\}\\
</task>\\

Every round I will give you an observation and a list of available actions, you have to respond an action based on the state and instruction. You can use search action if search is available. You can click one of the buttons in clickables.\\

The available actions are:\\
    \hspace*{1em} click[value]: click a button\\
    \hspace*{1em} search[keywords]: search for a keyword\\

If the action is not valid, perform nothing. Keywords in search are up to you, but the value in click must be a value in the list of available actions. Remember that your keywords in search should be carefully designed.\\

Below is the standard and detailed procedure for solving this task:\\
<conversation>\\
\{conversation\}\\
</conversation>\\

The generated meta plan should be written in the following format:\\
<meta\_plan>\\
Step 1: ...\\
Step 2: ...\\
...\\
</meta\_plan>\\
\end{tcolorbox}
\begin{figure*}[ht]
    \centering
    \vspace{-8pt}
    \caption{
    Prompt for WebShop Meta Plan Collection.
    }
    \label{fig: webshop meta plan prompt}
\end{figure*}

\begin{tcolorbox}[breakable, title=Instruction Prompt for ScienceWorld, enhanced jigsaw]
You are a helpful assistant to do some scientific experiment in an environment.\\
In the environment, there are several rooms: kitchen, foundry, workshop, bathroom, outside, living room, bedroom, greenhouse, art studio, hallway.\\
You should explore the environment and find the items you need to complete the experiment.
You can teleport to any room in one step.\\
All containers in the environment have already been opened, you can directly get items from the containers.\\
For each of your turn, you will be given the observation of the last turn. You should choose from two actions: "Thought" or "Action". If you choose "Thought", you should first think about the current condition and plan for your future actions, and then output your action in this turn. Your output must strictly follow this format:"Thought: your thoughts.\textbackslash n Action: your next action"; If you choose "Action", you should directly output the action in this turn. Your output must strictly follow this format:"Action: your next action". Remember that you can only output one "Action:" in per response.\\

The available actions are:\\
\hspace*{1em} open OBJ: open a container\\
\hspace*{1em} close OBJ: close a container\\
\hspace*{1em} activate OBJ: activate a device\\
\hspace*{1em} deactivate OBJ: deactivate a device\\
\hspace*{1em} connect OBJ to OBJ: connect electrical components\\
\hspace*{1em} disconnect OBJ: disconnect electrical components\\
\hspace*{1em} use OBJ [on OBJ]: use a device/item\\
\hspace*{1em} look around: describe the current room\\
\hspace*{1em} examine OBJ: describe an object in detail\\
\hspace*{1em} look at OBJ: describe a container's contents\\
\hspace*{1em} read OBJ: read a note or book\\
\hspace*{1em} move OBJ to OBJ: move an object to a container\\
\hspace*{1em} pick up OBJ: move an object to the inventory\\
\hspace*{1em} pour OBJ into OBJ: pour a liquid into a container\\
\hspace*{1em} mix OBJ: chemically mix a container\\
\hspace*{1em} teleport to LOC: teleport to a specific room\\
\hspace*{1em} focus on OBJ: signal intent on a task object\\
\hspace*{1em} wait: task no action for 10 steps\\
\hspace*{1em} wait1: task no action for a step

- - -\\
Here is an example.\\
\{example\}\\
- - -\\

Now, it's your turn and here is the task.\\
\{task\_instruction\}\\

This meta plan maybe helpful for you to complete the task:\\
\{meta\_plan\}
\end{tcolorbox}
\begin{figure*}[ht]
    \centering
    \vspace{-8pt}
    \caption{
    Instruction prompt for ScienceWorld.
    }
    \label{fig: scienceworld prompt}
\end{figure*}

\begin{tcolorbox}[breakable, title=Instruction Prompt for ALFWorld, enhanced jigsaw]
Interact with a household to solve a task. Imagine you are an intelligent agent in a household environment and your target is to perform actions to complete the task goal. At the beginning of your interactions, you will be given the detailed description of the current environment and your goal to accomplish. \\
For each of your turn, you will be given the observation of the last turn. You should choose from two actions: "Thought" or "Action". If you choose "Thought", you should first think about the current condition and plan for your future actions, and then output your action in this turn. Your output must strictly follow this format:"Thought: your thoughts.\textbackslash n Action: your next action"; If you choose "Action", you should directly output the action in this turn. Your output must strictly follow this format:"Action: your next action". \\
The available actions are:\\
\hspace*{1em} 1. go to {recep}\\
\hspace*{1em} 2. take {obj} from {recep}\\
\hspace*{1em} 3. put {obj} in/on {recep}\\
\hspace*{1em} 4. open {recep}\\
\hspace*{1em} 5. close {recep}\\
\hspace*{1em} 6. toggle {obj} {recep}\\
\hspace*{1em} 7. clean {obj} with {recep}\\
\hspace*{1em} 8. heat {obj} with {recep}\\
\hspace*{1em} 9. cool {obj} with {recep}\\
where {obj} and {recep} correspond to objects and receptacles.\\
After your each turn, the environment will give you immediate feedback based on which you plan your next few steps. if the envrionment output "Nothing happened", that means the previous action is invalid and you should try more options.\\
Reminder: \\
1. The action must be chosen from the given available actions. Any actions except provided available actions will be regarded as illegal.\\
2. Think when necessary, try to act directly more in the process.\\

- - -\\
Here is an example.\\
\{example\}\\
- - -\\

Now, it's your turn and here is the task.\\
\{task\_instruction\}\\

This meta plan maybe helpful for you to complete the task:\\
\{meta\_plan\}
\end{tcolorbox}
\begin{figure*}[ht]
    \centering
    \vspace{-8pt}
    \caption{
    Instruction prompt for ALFWorld.
    }
    \label{fig: alfworld prompt}
\end{figure*}

\begin{tcolorbox}[breakable, title=Instruction Prompt for WebShop, enhanced jigsaw]
You are web shopping. I will give you instructions about what to do. You have to follow the instructions.\\
Every round I will give you an observation and a list of available actions, you have to respond an action based on the state and instruction. You can use search action if search is available. You can click one of the buttons in clickables.\\
An action should be of the following structure:\\
\hspace*{1em} search[keywords] \\
\hspace*{1em} click[value]\\
If the action is not valid, perform nothing. Keywords in search are up to you, but the value in click must be a value in the list of available actions. Remember that your keywords in search should be carefully designed. \\
Your response should use the following format:\\
Thought: I think ...\\
Action: click[something]\\

- - -\\
Here is an example.\\
\{example\}\\
- - -\\

Now, it's your turn and here is the task.\\
\{task\_instruction\}\\

This meta plan maybe helpful for you to complete the task:\\
\{meta\_plan\}
\end{tcolorbox}
\begin{figure*}[ht]
    \centering
    \vspace{-8pt}
    \caption{
    Instruction prompt for WebShop.
    }
    \label{fig: webshop prompt}
\end{figure*}

\begin{tcolorbox}[breakable, title=Instruction Prompt for GPT Automated Assessment, enhanced jigsaw]
Please act as a professional instruction evaluator and assess the following two sets of meta plans.\\

Task description: \{task\}\\

DPO Plan:\\
\{dpo\}\\

SFT Plan:\\
\{sft\}\\

Please compare these two sets of meta plans across the following three dimensions:\\
1. Correctness - Does the meta plan accurately fulfill the task requirements?\\
2. Followability - Is the meta plan clear, easy to understand, and are the steps reasonable?\\
3. Standardization - Does the meta plan follow a consistent and standardized format?\\

For each dimension, please indicate which meta plan is better and provide reasoning. Finally, provide an overall assessment.\\
Please output the result in JSON format, including the following fields:\\
\{\\
    \hspace*{1em} "correctness\_better": "dpo"/"sft"/"tie",\\
    \hspace*{1em} "correctness\_reason": "reason",\\
    \hspace*{1em} "followability\_better": "dpo"/"sft"/"tie",\\
    \hspace*{1em} "followability\_reason": "reason",\\
    \hspace*{1em} "standardization\_better": "dpo"/"sft"/"tie",\\
    \hspace*{1em} "standardization\_reason": "reason",\\
    \hspace*{1em} "overall\_better": "dpo"/"sft"/"tie"\\
\}

\end{tcolorbox}
\begin{figure*}[ht]
    \centering
    \vspace{-8pt}
    \caption{
    Instruction prompt for GPT automated assessment.
    }
    \label{fig: prompt for gpt automated assessment}
\end{figure*}

\tcbset{width=\textwidth}
\begin{tcolorbox}[breakable, title=Case Study of ALFWorld, enhanced jigsaw]
\columnseprule=0.5pt
\begin{multicols}{2}
\begin{center}\textcolor{red}{\textbf{\Large SFT}}\end{center}
\textcolor{case_purple}{\textbf{Task Instruction:}} You are in the middle of a room. Looking quickly around you, you see a armchair 1, a cabinet 4, a cabinet 3, a cabinet 2, a cabinet 1, a drawer 5, a drawer 4, a drawer 3, a drawer 2, a drawer 1, a dresser 1, a garbagecan 1, a safe 1, a shelf 12, a shelf 11, a shelf 10, a shelf 9, a shelf 8, a shelf 7, a shelf 6, a shelf 5, a shelf 4, a shelf 3, a shelf 2, a shelf 1, a sidetable 1, and a sofa 1. Your task is to: find two pillow and put them in sofa.\\
\sethlcolor{red!50}\hl{\textit{\# Bad Meta Plan}}\\
\textcolor{mintleaf}{Meta Plan}: Step 1: go to sofa. \sethlcolor{red!50}\hl{Step 2: go to sidetable.} Step 3: take pillow from sidetable. Step 4: go to sofa. Step 5: put pillow in/on sofa. \sethlcolor{red!50}\hl{Step 6: go to sidetable.} Step 7: take another pillow from sidetable. Step 8: go to sofa. Step 9: put second pillow in/on sofa.\\
\textcolor{case_blue}{\textbf{Thought:}} To solve the task, I need to find two pillow and put them in sofa. \sethlcolor{red!50}\hl{The meta plan suggests that I should go to sofa first}, then go to sidetable.\\
\textcolor{case_blue}{\textbf{Action:}} go to sofa\\
\textcolor{case_purple}{\textbf{Observation:}} Nothing happens.\\
\textcolor{case_blue}{\textbf{Thought:}} It seems that the action "go to sofa" is invalid. \sethlcolor{red!50}\hl{I can try to go to sofa again, or I can try to go to sidetable first.}\\
\textcolor{case_blue}{\textbf{Action:}} go to sidetable\\
\textcolor{case_purple}{\textbf{Observation:}} Nothing happens.\\
\textcolor{case_blue}{\textbf{Thought:}} It seems that the action "go to sidetable" is also invalid. I can try to go to sofa again, or I can try to go to a cabinet or a drawer instead.\\
\textcolor{case_blue}{\textbf{Action:}} go to cabinet 1\\
\textcolor{case_purple}{\textbf{Observation:}} The cabinet 1 is closed.\\
\textcolor{case_blue}{\textbf{Thought:}} Now that I am in front of the cabinet 1, I can try to open it to see if there are any pillow inside.\\
\textcolor{case_blue}{\textbf{Action:}} open cabinet 1\\
\textcolor{case_purple}{\textbf{Observation:}} You open the cabinet 1. The cabinet 1 is open. In it, you see a tissuebox 1.\\
\textcolor{case_blue}{\textbf{Thought:}} It seems that there is no pillow in the cabinet 1. \sethlcolor{red!50}\hl{I can try to go to sidetable again to see if there is a pillow.}\\
\textcolor{case_blue}{\textbf{Action:}} go to sidetable\\
\textcolor{case_purple}{\textbf{Observation:}} Nothing happens.\\
\columnbreak

\begin{center}\textcolor{my_green}{\textbf{\Large MPO}}\end{center}
\textcolor{case_purple}{\textbf{Task Instruction:}} You are in the middle of a room. Looking quickly around you, you see a armchair 1, a cabinet 4, a cabinet 3, a cabinet 2, a cabinet 1, a drawer 5, a drawer 4, a drawer 3, a drawer 2, a drawer 1, a dresser 1, a garbagecan 1, a safe 1, a shelf 12, a shelf 11, a shelf 10, a shelf 9, a shelf 8, a shelf 7, a shelf 6, a shelf 5, a shelf 4, a shelf 3, a shelf 2, a shelf 1, a sidetable 1, and a sofa 1. Your task is to: find two pillow and put them in sofa.\\
\sethlcolor{green!50}\hl{\textit{\# Good Meta Plan}}\\
\textcolor{mintleaf}{Meta Plan}: \sethlcolor{green!50}\hl{Step 1: go to where the first pillow may be located.} Step 2: take first pillow. Step 3: go to where sofa is. Step 4: put first pillow in/on sofa. Step 5: go to where the second pillow may be located. Step 6: take second pillow. Step 7: go back to sofa. Step 8: put second pillow in/on sofa.\\
\textcolor{case_blue}{\textbf{Thought:}}  A pillow is more likely to appear in armchair (1), cabinet (1-4), drawer (1-5), dresser (1), shelf (1-12), sofa (1). \sethlcolor{green!50}\hl{I can check one by one, starting with armchair 1.}\\
\textcolor{case_blue}{\textbf{Action:}} go to armchair 1\\
\textcolor{case_purple}{\textbf{Observation:}} On the armchair 1, you see a pillow 2, a pillow 1, and a remotecontrol 1.\\
\textcolor{case_blue}{\textbf{Thought:}} Now I find the first pillow (1). Next, I need to take it. I can find the second pillow (2) later in armchair 1.\\
\textcolor{case_blue}{\textbf{Action:}} take pillow 1 from armchair 1\\
\textcolor{case_purple}{\textbf{Observation:}} You pick up the pillow 1 from the armchair 1.\\
\textcolor{case_blue}{\textbf{Thought:}} Now I take the first pillow (1). \sethlcolor{green!50}\hl{Next, I need to put it in/on sofa. I can directly go to sofa.}\\
\textcolor{case_blue}{\textbf{Action:}} go to sofa 1\\
\textcolor{case_purple}{\textbf{Observation:}} On the sofa 1, you see nothing.\\
\textcolor{case_blue}{\textbf{Thought:}} Now I go to sofa. Next, I need to put the first pillow in/on sofa. I can directly put the first pillow in/on sofa.\\
\textcolor{case_blue}{\textbf{Action:}} put pillow 1 in/on sofa 1\\
\textcolor{case_purple}{\textbf{Observation:}} You put the pillow 1 in/on the sofa 1.\\
\textcolor{case_blue}{\textbf{Thought:}} Now I put the first pillow in sofa. Next, I need to find the second pillow. I can directly go to armchair 1.\\
\end{multicols}

\begin{multicols}{2}
\textcolor{case_blue}{\textbf{Thought:}} It seems that the action "go to sidetable" is still invalid. I can try to go to sofa again to start the workflow.\\
\textcolor{case_blue}{\textbf{Action:}} go to sofa\\
\textcolor{case_purple}{\textbf{Observation:}} Nothing happens.\\
\textcolor{case_blue}{\textbf{Thought:}} It seems that the action "go to sofa" is still invalid. I can try to go to a drawer instead, as the workflow does not specify which drawer to go to.\\
\textcolor{case_blue}{\textbf{Action:}} go to drawer 1\\
\textcolor{case_purple}{\textbf{Observation:}} The drawer 1 is closed.\\
....\\
....\\
....\\
\textcolor{case_blue}{\textbf{Thought:}} It seems that there is no pillow in the cabinet 2. I can try to go to a different cabinet or location.\\
\textcolor{case_blue}{\textbf{Action:}} go to cabinet 3\\
\textcolor{case_purple}{\textbf{Observation:}} The cabinet 3 is closed.\\
\textcolor{case_blue}{\textbf{Thought:}} Now that I am in front of the cabinet 3, I can try to open it to see if there are any pillow inside.\\
\textcolor{case_blue}{\textbf{Action:}} open cabinet 3\\
\textcolor{case_purple}{\textbf{Observation:}} You open the cabinet 3. The cabinet 3 is open. In it, you see a box 3, and a tissuebox 2.\\
\textcolor{case_purple}{\textbf{Reward:}} 0.0\\
\columnbreak

\textcolor{case_blue}{\textbf{Action:}} go to armchair 1\\
\textcolor{case_purple}{\textbf{Observation:}} On the armchair 1, you see a pillow 2, and a remotecontrol 1.\\
\textcolor{case_blue}{\textbf{Thought:}} Now I find the second pillow (2). Next, I need to take it.\\
\textcolor{case_blue}{\textbf{Action:}} take pillow 2 from armchair 1\\
\textcolor{case_purple}{\textbf{Observation:}} You pick up the pillow 2 from the armchair 1.\\
\textcolor{case_blue}{\textbf{Thought:}} Now I take the second pillow (2). Next, I need to put it in/on sofa.\\
\textcolor{case_blue}{\textbf{Action:}} go to sofa 1\\
\textcolor{case_purple}{\textbf{Observation:}} On the sofa 1, you see a pillow 1.\\
\textcolor{case_blue}{\textbf{Thought:}} Now I go to sofa. Next, I need to put the second pillow in/on sofa. I can directly put the second pillow in/on sofa.\\
\textcolor{case_blue}{\textbf{Action:}} put pillow 2 in/on sofa 1\\
\textcolor{case_purple}{\textbf{Observation:}} You put the pillow 2 in/on the sofa 1.\\
\textcolor{case_purple}{\textbf{Reward:}} 1.0\\

\end{multicols}

\end{tcolorbox}
\begin{figure}[!h]
    \centering
    \vspace{-8pt}
    \caption{
    Case study of ALFWorld.
    }
    \label{fig:alfworld_case}
\end{figure}

\end{document}